\documentclass[letterpaper, 10 pt, conference]{ieeeconf}  

\IEEEoverridecommandlockouts
\usepackage{amsmath}
\usepackage{amsfonts}
\usepackage{amssymb}
\usepackage{multirow}
\usepackage{colortbl}
\usepackage{graphicx}
\usepackage{subfigure}
\usepackage{array}
\usepackage{algorithm}
\usepackage{algorithmic}
\usepackage{hhline}
\usepackage{color}
\usepackage{xspace}
\usepackage{ifpdf}

\ifpdf 
  \DeclareGraphicsExtensions{.pdf,.png,.jpg,.jpeg}
\else 
  \DeclareGraphicsExtensions{.eps,.bmp}
\fi

\DeclareMathOperator*{\argmin}{arg\,min}

\makeatletter
\DeclareRobustCommand\onedot{\futurelet\@let@token\@onedot}
\def\@onedot{\ifx\@let@token.\else.\null\fi\xspace}

\def\ie{\emph{i.e}\onedot}

\def\eg{\emph{e.g}\onedot}

\def\etal{\emph{et al}\onedot}

\def\wrt{w.r.t\onedot}
\def\thresh{T}
\def\trsp{\mathsf{T}}
\def\cov{\Sigma}
\def\err{\mathbf{e}}
\def\cost{\rho}

\def\poses{\mathcal{X}}
\def\points{Y}
\def\inlier_ratio{\gamma}
\DeclareMathOperator{\prob}{prob}

\title{\LARGE \bf
Noise Models in Feature-based Stereo Visual Odometry
}

\author{Pablo F. Alcantarilla$^\dag$ and Oliver J. Woodford$^\ddag$
\thanks{This work was done while both authors were employees at Toshiba Research Europe Ltd., Cambridge, United Kingdom. Contact information: {\tt 
pablofdezalc@gmail.com}$^\dag$, {\tt o.j.woodford.98@cantab.net}$^\ddag$.}%
}

\begin{document}

\maketitle
\thispagestyle{empty}
\pagestyle{empty}


\begin{abstract}
Feature-based visual structure and motion reconstruction pipelines, common in visual odometry and large-scale 
reconstruction from photos, use the location of corresponding features in different images to determine the 3D structure 
of the scene, as well as the camera parameters associated with each image. The noise model, which defines the likelihood 
of the location of each feature in each image, is a key factor in the accuracy of such pipelines, alongside optimization 
strategy. Many different noise models have been proposed in the literature; in this paper we investigate the performance 
of several. We evaluate these models specifically \wrt stereo visual odometry, as this task is both simple (camera 
intrinsics are constant and known; geometry can be initialized reliably) and has datasets with ground truth readily 
available (KITTI Odometry and New Tsukuba Stereo Dataset). Our evaluation shows that noise models which are more 
adaptable to the varying nature of noise generally perform better.
\end{abstract}

%
\section{Introduction}\label{sec:introduction}
Inverse problems---given a set of observed measurements, infer the physical system that generated them---are common in computer vision and 
robotics. One of the keys to solving such problems is knowing how accurate each measurement is. This is captured in a noise model: a 
distribution over expected measured values, given a true value.

In visual reconstruction tasks the measurements are images and the physical system is scene geometry, lighting and camera position. 
\emph{Direct} methods~\cite{Comport10ijrr,Lovegrove11iv} phrase the problem in exactly this way, seeking the scene/camera model which best 
recreates all the pixel values of the images. The source of noise in this case is well understood: mostly electronic noise (for electronic 
sensors). It is also straightforward to measure. However, most direct methods simply approximate the noise with a Gaussian or Laplacian, and 
this seems to be sufficient.

\begin{figure}[ht]
\centering
\includegraphics[width=0.48\textwidth]{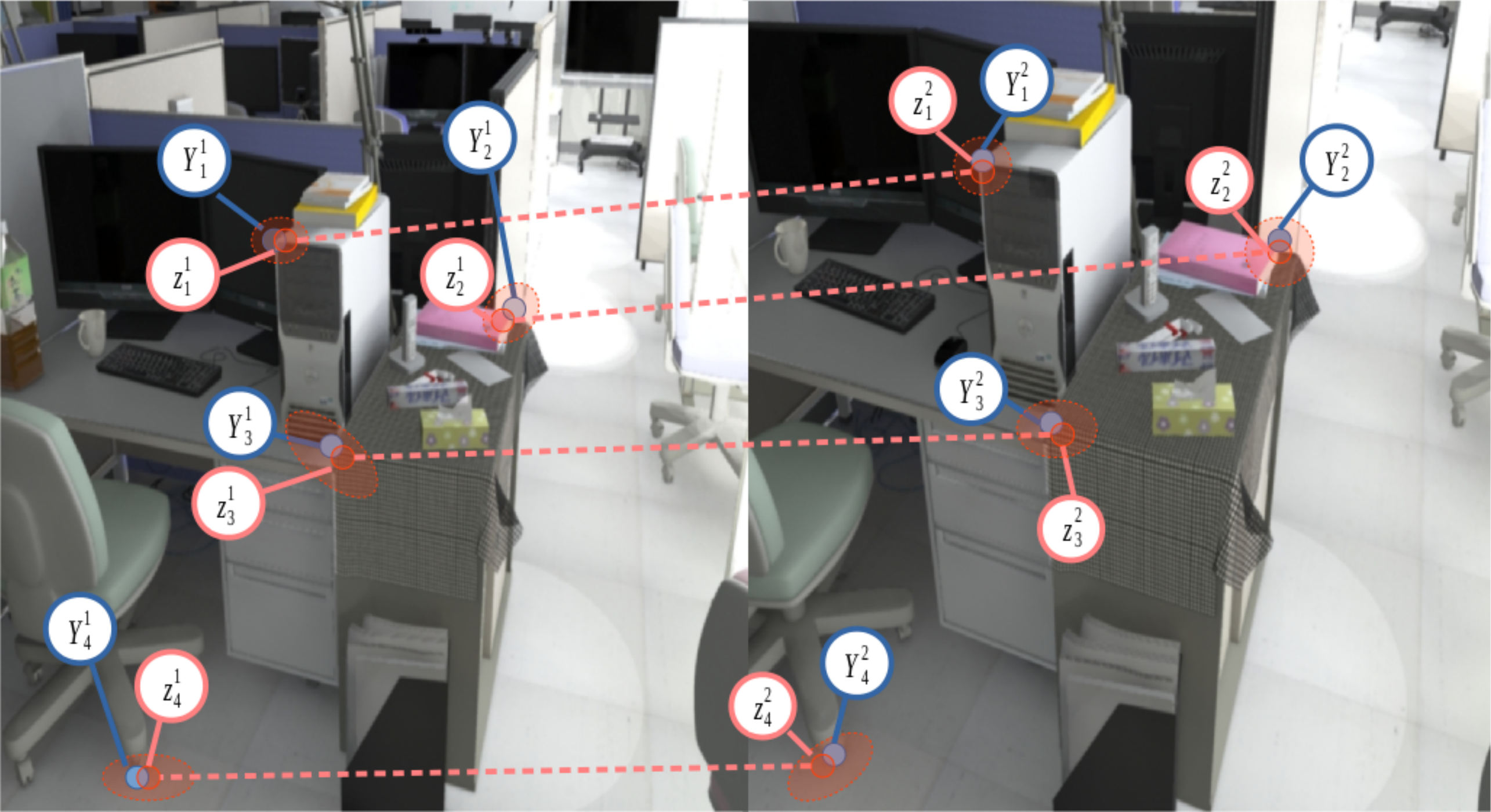}\\
\caption{Given a set of 2D feature correspondences $z_i$ (red circles) between two images, the inverse problem to be solved is to 
estimate the camera poses and the 3D locations of the features $Y_i$. The true 3D locations project onto each image plane to give a true 
image location (blue circles). The noise model represents a probability distribution over the error $\err$ between the true and measured 
feature locations. The size of the ellipses depends on the uncertainty in the feature locations.}\label{fig:stereo_vo}
\end{figure}

In contrast, \emph{feature-based} methods~\cite{Beall10iros,Geiger11iv,Badino13cvad} pre-process the images, extracting a set of interest 
points, or features, from each and tracking or matching these across images. In this case, the inverse problem to be solved is: Given 
feature image locations, compute their 3D locations and the camera poses. This gives rise to the following notion of a \emph{feature noise 
model}, illustrated in Fig.~\ref{fig:stereo_vo}: Features are assumed to be points in space, the true 3D locations of which are unknown 
variables. These features project onto each image plane to give a true image location. The ``measurements'' are the computed image locations 
of these features, and are a noisy version of the true values. The noise model is therefore a probability distribution over the expected 
image location of a measurement, given the true image location.

In this case, the nature of the noise distribution is not so obvious. It is certainly not well approximated by a Gaussian, since errors in 
tracking or matching can lead to gross errors, or outliers.

Several noise models, along with associated optimization strategies, have been suggested in the literature, which we discuss in detail in 
$\S\ref{sec:noise_models}$. However, there is currently lacking an up-to-date evaluation and comparison of these noise models. Previous 
approaches such as~\cite{Choi09bmvc}, evaluate the performance of RANSAC-based~\cite{Bolles81ijcai} noise models for line 
fitting and planar homography estimation. Recently in~\cite{Liu14eccv}, a match selection and refinement strategy for two-view Structure 
from Motion~(SfM) problems was presented. That work analyzes the trade-offs of quality versus quantity of feature matches applied to 
different noise models.

In this work, we present a thorough explanation of feature noise models and unify the explanation to describe how each one specifies a 
given distribution on the measurement errors. We evaluate published feature noise models along with optimization strategies applied to the 
problem of feature-based stereo visual odometry (VO)~\cite{Nister04cvpr,Scaramuzza11ram}. This is a simple visual reconstruction task; 
camera intrinsics are known and constant, stereo gives a good initial structure estimate, and the baseline between frames is small. It also 
has good test datasets and performance benchmarks available, in the form of the KITTI Odometry benchmark~\cite{Geiger13ijrr} and the New 
Tsukuba Stereo dataset (synthetic)~\cite{Peris12icpr}.

The next section provides an overview of noise models used in feature-based visual reconstruction, focussing on those we evaluate and 
applied to the problem of stereo VO. In $\S\ref{sec:stereo_vo}$ we describe our evaluation framework. In $\S\ref{sec:results}$ we present 
the results of our evaluation, before concluding in the final section.

\section{Feature noise models} \label{sec:noise_models}
In this section we review feature noise models in the literature. Many visual reconstruction methods do not in fact mention probability 
distributions. Some talk about cost functions and optimizations; others simply present algorithms. However, all these methods do imply 
assumptions about the nature of the noise. In order to present these assumptions on an even field, we cast them all as explicit probability 
distributions.

To simplify things, we refer directly to the difference in image coordinates between the true and measured feature locations, which we 
denote as $\err$ (see Fig. \ref{fig:stereo_vo}). We therefore assume that our reader is familiar with the projection of 3D points in world 
coordinates onto image planes \cite{Hartley04book}. Since many noise models are defined by summed costs, \ie the negative log likelihood of 
probability, we represent such cost functions by $\rho(\err)$, and define a helper function,
\begin{equation}\label{eq:probability}
\prob\left(\cost(\err)\right) = \frac{\exp\left(-\cost(\err)\right)}{\int\exp\left(-\cost(\mathbf{x})\right) \mathrm{d} \mathbf{x}},
\end{equation}
to convert a cost to a probability.

The set of variables, $\Theta$, used to compute the measurement errors, $\{\err_i\}_{i=1}^N$, consists of camera pose variables (extrinsics) $\poses$, 3D feature locations $\points$, and optionally camera intrinsics and noise model parameters.
The cost, which is the negative log of the probability of the measurements, for a given $\Theta$ is defined as
\begin{equation}\label{eq:cost}
E(\Theta) = \underbrace{\sum\limits_{i=1}^{N} \rho\Big(\err_{i}\Big)}_{\textrm{Data costs}} - \underbrace{N\cdot\log\left(\int\exp\left(-\cost(\mathbf{x})\right) \mathrm{d} \mathbf{x}\right)}_{\textrm{Noise model cost (usually constant)}}.
\end{equation}
Two aspects of feature-based visual reconstruction make the above cost multi-modal, irrespective of the noise models used. 
Firstly, the projection of features onto image planes and the camera rotations are both non-linear operations; convex noise models do not 
therefore lead to a convex cost \wrt the variables. Secondly, the presence of outliers encourage spurious local minima.

Most visual reconstruction pipelines therefore have two distinct stages (following the feature extraction stage): An 
\emph{initialization} stage, where approximate 3D feature locations and camera poses are computed. The main tasks here are to efficiently 
find a starting point in the convergence basin of the global minimum of the cost function, and to reject 
outlier measurements in the process. Then a \emph{refinement} stage, where the remaining measurements are used to iteratively improve output 
variables. The main task here is to maximize the accuracy of the final solution.

Since the noise models of the first stage need to be robust to outliers, whilst those of the second stage need to maximize accuracy of 
inliers, the models used in each stage are often different. We therefore split our discussion of models into those two stages.

\subsection{Initialization}\label{sec:noise_init}
The majority of initialization methods in visual reconstruction pipelines use a \emph{hypothesize-and-test} framework 
to select an initial set of variables $\Theta$: many sets of values for the variables are computed,\!\!\footnote{The 
method for computing plausible sets of values is application specific. We describe our approach for stereo VO in 
$\S\ref{sec:stereo_vo}$.} and the set with the lowest cost (computed using Eq.~(\ref{eq:cost})) is selected. This 
helps to ensure the initial values lie within the convergence basin of the global minimum. The noise models below are 
all used within this framework.

\subsubsection{\textbf{RANSAC}}
\emph{RAndom SAmple Consensus}~\cite{Bolles81ijcai} is the original hypothesize-and-test framework. Its score is the number of errors below 
a given threshold, $\thresh$.
This corresponds to the following cost function:
\begin{equation}\label{eq:ransac_rho}
\rho(\err) = \begin{cases} 0 & \mbox{if } \|\err\|^2 < \thresh^2, \\
1 & \mbox{otherwise}. \end{cases}
\end{equation}
The threshold $\thresh$ quantifies the maximum deviation attributable to the effects of noise and is used to classify each correspondence 
between inlier or outlier. Both are assumed to be distributed uniformly within their domains. 

The choice of this threshold $\thresh$ has a tremendous impact on the accuracy of the estimated variables. If $\thresh$ 
is too small, little data is used to estimate the variables, which can lead to inaccuracies; on the other hand, if 
$\thresh$ is too large, the estimated variables may be corrupted by outliers.

\subsubsection{\textbf{MSAC}}
Torr \etal~\cite{Torr00cviu} recognised that a uniform distribution is a poor approximation for inliers. They instead proposed the MSAC 
cost function, which is a slight variation on RANSAC that assumes that inliers are distributed normally, thus:
\begin{equation}\label{eq:msac_rho}
\rho(\err) = \begin{cases} \|\err\|^2 & \mbox{if } \|\err\|^2 < \thresh^2, \\
\thresh^2 & \mbox{otherwise}. \end{cases}
\end{equation}

\subsubsection{\textbf{MLESAC}}
In the same paper, Torr \etal~\cite{Torr00cviu} take a more probabilistic approach with their maximum likelihood estimator, MLESAC. 
Similarly to MSAC, it assumes that inliers are distributed normally, though with a full covariance matrix, and that outliers are distributed 
uniformly. However, rather than assuming (or implying) a prior probability for a measurement being an inlier, they introduce a new variable, 
$\inlier_ratio$, to represent this. This leads to the following noise model:
\begin{equation}\label{eq:mlesac}
\rho(\err) = -\log\left[\inlier_ratio\prob\left(\frac{1}{2} \err^\trsp \cov^{-1} \err \right) + 
(1-\gamma)\frac{1}{\nu}\right]
\end{equation}
where $\nu$ is the space within which the outliers are believed to 
fall uniformly and $\cov$ is the covariance matrix of the inlier noise. For each hypothesis, the value of $\inlier_ratio$ that maximizes 
likelihood is computed using Expectation-Maximization (EM) from some initial value, \eg $0.5$. The benefit of this approach is that it can 
adapt to varying levels of outlier contamination.

\subsubsection{\textbf{AMLESAC}}
MLESAC assumes prior knowledge of the inlier noise distribution, through the parameter $\cov$. In certain scenarios this distribution may 
not be known, or may even vary according to the conditions. To overcome this, Adaptive MLESAC~\cite{Konouchine05graphicon} has the same 
noise model as Eq.~(\ref{eq:mlesac}), but additionally optimizes $\cov$.

\subsubsection{\textbf{AC-RANSAC}}\label{sec:acransac}
AC-RANSAC~\cite{Moisan04ijcv,Moulon12accv} uses the \emph{a contrario} methodology. For each hypothesis, this approach computes an 
inlier/outlier adaptive threshold based on the \emph{Helmholtz principle}---whenever some large deviation from randomness occurs, a 
structure is perceived. AC-RANSAC seeks to minimize the number of false alarms~(\textit{NFA}). 
\begin{equation} \label{eq:ac_ransac_cost}
\hat{\Theta} \ = \ \argmin_{\Theta} \ \textit{NFA}(\err, q) \leq \epsilon
\end{equation}
where $\epsilon$ is the minimum number of false alarms in order to consider a set of variables $\Theta$ as valid and it is usually set to 
$1$. A false alarm is defined as a set of variables that is actually due to chance. This requires the definition of a 
rigidity measure $\err$ and a background noise model $\mathcal{H}_0$, that models random correspondences assuming that 
points are uniformly distributed in their respective images. The \textit{NFA} is defined as: 
\begin{equation} \label{eq:ac_ransac}
\begin{array}{c}
\textit{NFA}(\err, q) = (N-N_{s})\binom{N}{q}\binom{q}{N_{s}}\left(\err_{q}^{d} \alpha_{0} 
\right)^{q-N_{s}}
\end{array}
\end{equation}
where:
\begin{itemize}
\item $N$ is the total number of correspondences between two images.
\item $N_{s}$ is the cardinal of a RANSAC sample.
\item $q$ is the total number of hypothesized inlier correspondences.
\item $\err_{q}$ is the q-th lowest error among all $N$ correspondences and depends on the set of variables $\Theta$.
\item $d$ is the error dimension. 
\item $\alpha_{0}$ is the probability of a random correspondence of having error of one pixel. In the case of stereo VO, this probability 
is computed as the ratio between the volume 
\end{itemize}

In the particular case of stereo VO, $d=3$ for rectified stereo pairs and $\alpha_0$ can be computed as the ratio 
between the volume of a sphere of unit radius divided by the volume given by the image dimensions and the disparity 
range, \ie $4\pi/(3 \cdot \mathrm{Width} \cdot 
\mathrm{Height} \cdot \mathrm{Disparity})$.

\subsubsection{\textbf{ERODE}}
In contrast to the above noise models, the \textit{Efficient and Robust Outlier Detector}~\cite{Moreno13icra} is not used in a 
hypothesize-and-test framework. Instead, it uses a robust yet convex cost function, the Pseudo-Huber kernel, to reduce the number of local 
minima, and performs an optimization on the variables from a single starting point. The cost function is given by:
\begin{equation} \label{eq:pseudo_huber}
\cost(\err) = 2b^2 \left( \sqrt{1 + \frac{\|\err\|^2}{b^2}} - 1 \right)
\end{equation}
where $b$ is a user-defined parameter that tunes the shape of the function. ERODE is faster compared to RANSAC and other consensus 
approaches and achieves similar accuracy when the initial camera pose is close to the real one.

Other robust costs with adaptive thresholds have been proposed in the literature such as~\cite{Raguram11iccv} and more 
recently~\cite{Cohen15iccv}. These methods overcome the limitations of globally-fixed thresholds, yielding better precision. 
RECON~\cite{Raguram11iccv} is built on the observation that those sets of variables generated from uncontaminated minimal subsets of 
correspondences are consistent in terms of the behavior of their residuals, while contaminated subsets exhibit uncorrelated behavior. Cohen 
and Zach~\cite{Cohen15iccv} propose an improvement of RANSAC which also optimizes over a discrete number of outlier thresholds in a very 
efficient manner, using the likelihood-ratio test.

\begin{figure}[t]
\centering
\includegraphics[width=0.42\textwidth]{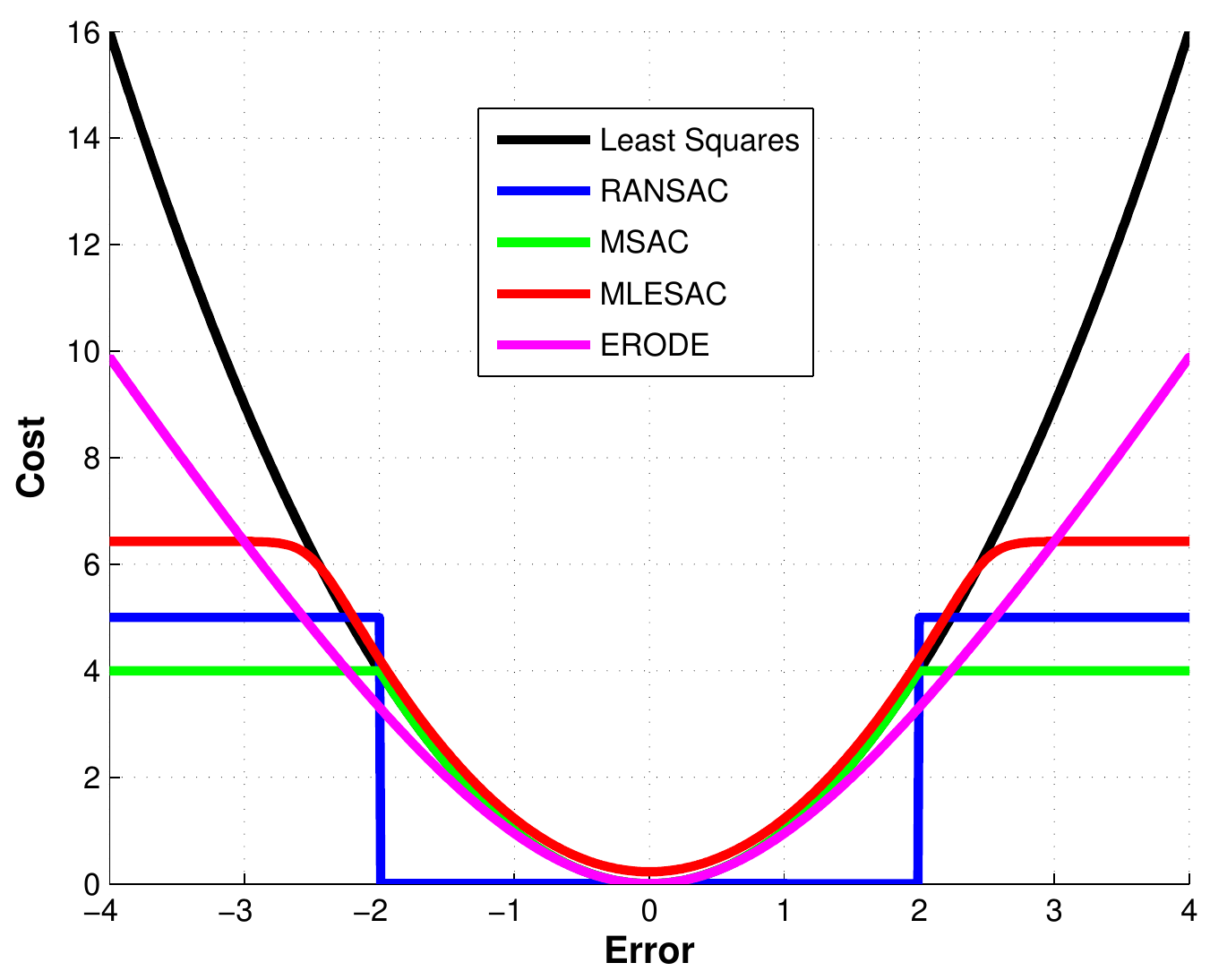}\\
\caption{Cost functions $\cost\left(\cdot \right)$ for several noise models: RANSAC, MSAC, MLESAC and ERODE with $b=2$. We also show the 
least squares cost function for illustrative purposes.}
\label{fig:cost_functions}
\end{figure}

Table~\ref{tab:initialization_methods} shows a summary of the robust initialization methods considered in this paper, 
showing their cost functions and main configuration parameters. Fig.~\ref{fig:cost_functions} depicts the cost 
functions for some of the noise models considered in this paper.
\begin{table}[htpb]
  \begin{small}
  \begin{center}
    \begin{tabular}{|c|c|c|c|c|}
      \hline
       & \textbf{Cost Function:}  & \textbf{Parameters}\\
      \hline
      \textbf{RANSAC} & Inliers Count & $T$\\
      \hline
      \textbf{MSAC} & Truncated Quadratic & $T$\\
      \hline
      \textbf{MLESAC} & Neg. Log Gaussian+Uniform &  $\Sigma$, $\nu$ \\
      \hline
      \textbf{ERODE} & Pseudo-Huber  & $b$, $T$\\
      \hline
      \textbf{AC-RANSAC} & Number of False Alarms & $\alpha_0, \epsilon$\\
      \hline
    \end{tabular}
  \end{center}
  \end{small}
  \caption{Robust initialization: cost functions and main configuration parameters.}\label{tab:initialization_methods}
\end{table}
%

\subsection{Refinement}\label{sec:noise_refine}
In the refinement stage, the goal is to maximize the accuracy of the final solution using a good initial set of variables. It is typically a minimization of the cost in equation (\ref{eq:cost}) using standard 
nonlinear least-squares optimizers such as Gauss-Newton or Levenberg-Marquardt.

\subsubsection{Handling outliers}
Three of the noise models from $\S\ref{sec:noise_init}$ robustly handle outliers, as well as defining a sensible noise model for inliers:  MSAC, MLESAC and ERODE. These can be used directly in a refinement optimization.
RANSAC's uniform inlier model does not lend itself to refinement, and AC-RANSAC does not define an inlier noise model. These approaches therefore need a different noise model for the refinement stage.
Both use the same strategy, which is to remove from the refinement optimization those measurements deteremined to be outliers during the initialization stage, \ie those measurements for which $\|e\|$ was above the outlier threshold for the selected $\Theta$.
The remaining measurements are then given a simple, uniform gaussian cost:
\begin{equation}
\rho(\err) = \|\err\|^2.
\end{equation}

\subsubsection{Variables optimized over}
The refinement optimization may not necessarily refine all the variables. As a minimum, the optimization refines camera pose variables, $\poses$, between pairs of frames. 3D structure, $\points$, can be added, creating a two-view Bundle Adjustment~(BA) 
optimization~\cite{Hartley04book,Triggs00}. It is also possible to include the parameters of the noise distribution (if there are any) as variables. Finally, more frames can be introduced into the refinement~\cite{Mouragnon09imavis,Indelman12bmvc}. Every additional variable comes at the 
expense of higher computational demand. In our experiments we will investigate three levels of variable refinement:
\begin{enumerate}
\item \textbf{Motion only}: The 3D structure $\points$ and noise parameters are fixed, and only camera poses $\poses$ are optimized.
\item \textbf{Motion and Structure (BA)}: The noise parameters are fixed, while camera poses $\poses$ and 3D structure $\points$ are optimized.
\item \textbf{Motion, Structure and Inlier Noise Distribution}: We change the inlier noise distribution from a uniform gaussian to a Cauchy distribution with full covariance:
\begin{equation}
\rho(\err) = -\log\left(1+\err^\trsp \cov^{-1} \err\right).
\end{equation}
Furthermore, the parameters of the inverse covariance,\!\!\footnote{We use a triangular parameterization of the square root of the inverse covariance matrix $\cov^{-1}$.} $\cov^{-1}$, are optimized in addition to the camera poses $\poses$ and the 3D structure $\points$.
Since the noise distribution is changing in the optimization, the right hand term of equation (\ref{eq:cost}) is not constant, and therefore should be included in the optimization; for the Cauchy distribution this term is
$-\log\left(|\cov|\right)\cdot N/(d+1)$.

Estimating the inlier noise distribution can help when this distribution is unknown or variable and non-uniform, \eg with occasional motion 
blur. We use the robust Cauchy distribution to add further protection against the presence of outliers.
\end{enumerate}

\section{Stereo Visual Odometry}\label{sec:stereo_vo}
The goal of stereo VO is to estimate the camera motion $\poses_{k}  = \{R,\mathbf{t}\}$ between two consecutive frames $\left(k-1, 
k\right)$ in the Euclidean group $SE(3)$ of 3D poses, by minimizing the reprojection error given a set of $N$ correspondences between 3D 
points $\points_{i} \in \mathbb{R}^{3}$ and stereo image measurements $z_{i} \in \mathbb{R}^{6}$. Stereo calibration parameters are obtained in a 
prior calibration process. The calibration parameters comprise of camera focal length $f$, principal point $\left(u_0, 
v_0\right)$, stereo baseline $B$ and distortion parameters. Given the stereo calibration parameters, \textit{stereo rectification} is 
typically performed to simplify the stereo correspondence problem.

Stereo VO has the benefit over monocular VO that an initial estimate of scene geometry, \ie 3D points, can be computed from the 2D 
correspondences from the stereo image pair at each timeframe, independently of any camera motion between frames. Current feature-based 
stereo VO approaches~\cite{Beall10iros,Geiger11iv,Badino13cvad} generally follow these three steps:
\begin{enumerate}
\item \textbf{Putative correspondences}: A tracking~\cite{Johnson08icra,Shi94icpr} or matching~\cite{Alcantarilla13bmvc,Lowe04ijcv} method 
establishes an initial set of putative 2D point correspondences between consecutive stereo frames. A single putative correspondence, 
comprises of a set of stereo image measurements $z_i = \left[u_{l}^{k-1}, u_{r}^{k-1}, v^{k-1}, u_{l}^{k}, u_{r}^{k}, v^{k}\right]^\trsp$ 
and triangulated 3D points $\points_i^{k-1}$ with respect to stereo frame $k-1$. The subscripts $l$ and $r$ denote measurements from the left and 
right images respectively.
\item \textbf{Robust initialization}: An inlier subset of the correspondences and an approximate initial camera motion and 3D scene 
geometry are computed.
\item \textbf{Model Refinement}: The camera motion, and optionally 3D scene geometry and inliers noise distribution, are refined by 
minimizing the 2D reprojection error of the inlier point correspondences in the image domain.
\end{enumerate}

In stereo VO, the error $\err$ between the predicted and measured image feature locations for a particular 
correspondence is defined as:
\begin{equation} \label{eq:error_svo}
\err_i \ = \ z_{i}^{k} - \pi \left(\poses_k \points_{i}^{k-1}\right)
\end{equation}
where the function $\pi(\cdot)$ is the stereo projection function that projects a 3D point from the stereo frame $k-1$ into the 
left and right images at frame $k$ given the stereo calibration parameters. In the case of rectified stereo, this error has 
dimension $d=3$. When refining also for the structure parameters, the error $\err$ takes into account the measurements in the two views 
$z_{i}^{k-1}$ and $z_{i}^{k}$, therefore having a dimension $d=6$. In this case, the camera pose for the frame $k-1$ is set to a 
canonical pose with identity rotation matrix and zero translation vector.

\section{Experimental Results}\label{sec:results}
In this section we show the results of our noise model evaluation \wrt feature-based stereo VO, alongside optimization 
strategies. First, we describe some details of our evaluation in~$\S\ref{sec:preliminaries}$. Then we show 
experimental results for the robust initialization and refinement steps in~$\S\ref{sec:robust_results}$ and 
$\S\ref{sec:refinement_results}$ respectively.

\subsection{Preliminaries}\label{sec:preliminaries}
In our evaluation we consider two stereo datasets: the KITTI odometry benchmark~\cite{Geiger13ijrr} and the New Tsukuba Stereo 
dataset~\cite{Peris12icpr}. The KITTI Odometry benchmark comprises 20 sequences captured from a front facing car mounted camera; ground truth pose information is available 
for the first 11 sequences. The New Tsukuba Stereo dataset is a synthetic dataset for the purpose of stereo matching and camera tracking evaluation. The 
dataset comprises 4 different illumination settings, from which we show average results on the 
\textit{fluorescent} and \textit{daylight} settings.

\subsubsection{\textbf{LIBVISO2}}
We compare our stereo visual odometry approaches with the LIBVISO2 library~\cite{Geiger11iv}. LIBVISO2 is a publicly available library that 
performs feature-based stereo VO, detecting corners and blobs from the input images and using horizontal and vertical Sobel filter 
responses as descriptors. An initial camera motion is estimated by means of a RANSAC approach using a globally-fixed threshold $T$ equal to 
$2$ pixels. The camera motion is then refined using only the inliers.

\subsubsection{\textbf{Feature Detection, Description and Matching}} 
In our experiments, we use A-KAZE~\cite{Alcantarilla13bmvc} as our feature detector and descriptor method due to its good detector 
repeatability and highly discriminative binary descriptors. We first match the binary descriptors between the left and right images of the 
stereo pair using a brute-force approach and consistency check between the two views to establish a set of \textit{stereo matches}. Then, 
for obtaining a set of temporal \textit{putative correspondences}, we perform the same matching strategy considering the left image as the 
reference one and again using a brute-force approach and consistency check between the two consecutive stereo frames. 

\subsubsection{\textbf{Initial Camera Motion}}
For each hypothesis in consesus-based methods, an initial motion is computed from a minimal set of $N_{s}$ correspondences. We obtain an 
initial guess for the camera motion by using a \textit{Perspective-n-Point}~(PnP) algorithm~\cite{Gao03pami} with $N_{s}=4$ points as 
minimal sample and considering only the measurements from the left camera. ERODE considers an initial camera pose with identity rotation 
matrix and zero translation vector, assuming that the camera motion between consecutive frames is small.

\subsubsection{\textbf{Inlier/Outlier Classification}}
We consider two different values of the threshold parameter $\thresh$ to highlight the sensitivity of the estimated camera motion with 
respect to the threshold parameter in RANSAC and MSAC. The first threshold is set to $\thresh = 2$ pixels. The second value comes from the 
cumulative chi-squared distribution assuming that the measurement error is Gaussian with zero mean and standard deviation $\sigma = 1$ 
pixels. Under this assumption, the value of the threshold parameter is $2.79$ pixels~\cite{Hartley04book}. 

\subsubsection{\textbf{Evaluation Criteria}}
We use the same evaluation method recommended by the KITTI dataset: we compute translational and rotational 
errors for all possible sub-sequences of length $\left(100,150,200,\ldots,800\right)$ meters and take the average. Since 
the travelled distance in the New Tsukuba Stereo dataset is smaller, we consider all possible sub-sequences of 
length $\left(5,10,50,100,150,\ldots,400\right)$ centimeters.

\subsection{Bias in the KITTI Odometry Benchmark}\label{sec:kitti_bias}
\begin{figure}[t]
\centering
\begin{tabular}{cc}
\includegraphics[width=0.22\textwidth]{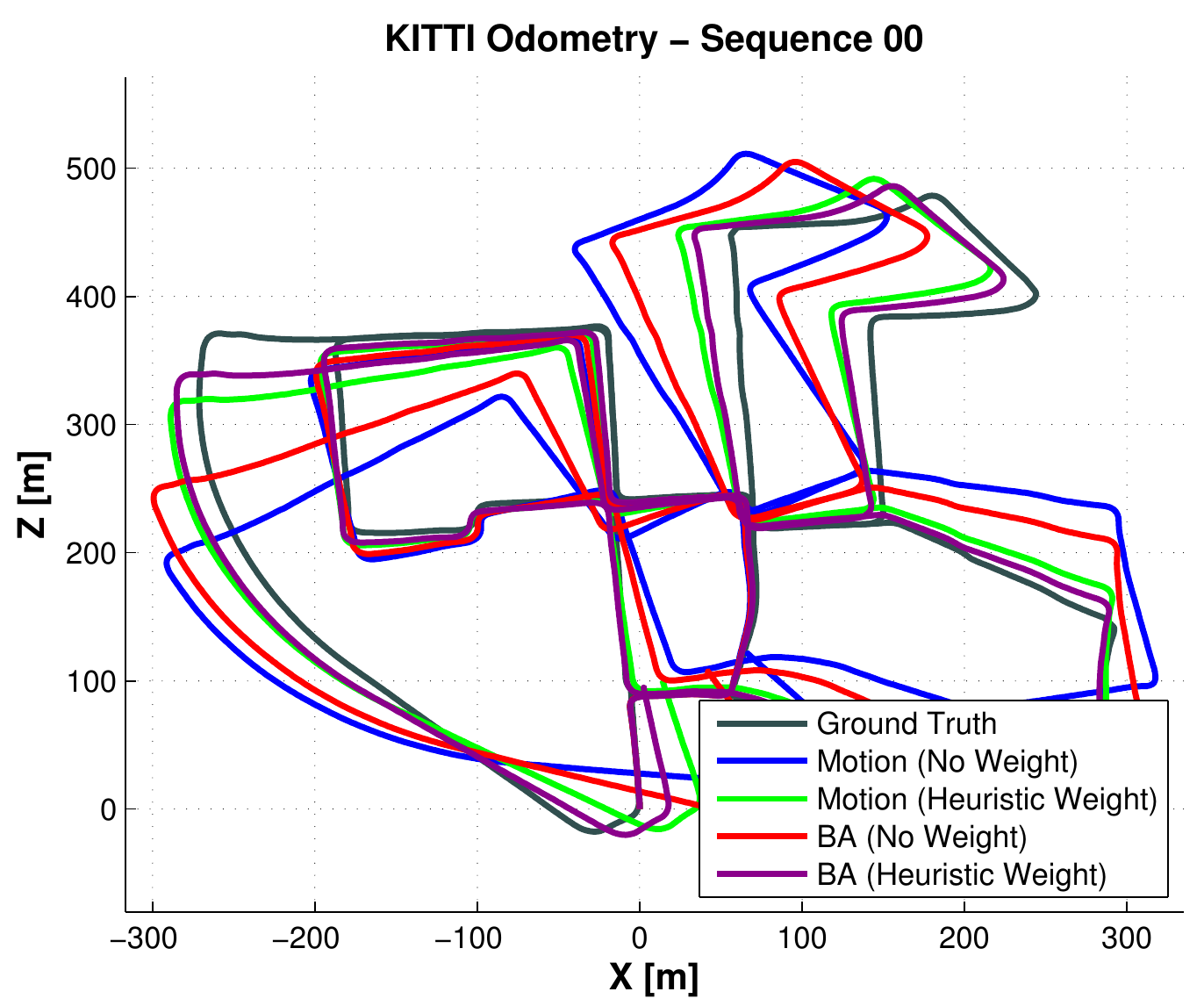}&
\includegraphics[width=0.22\textwidth]{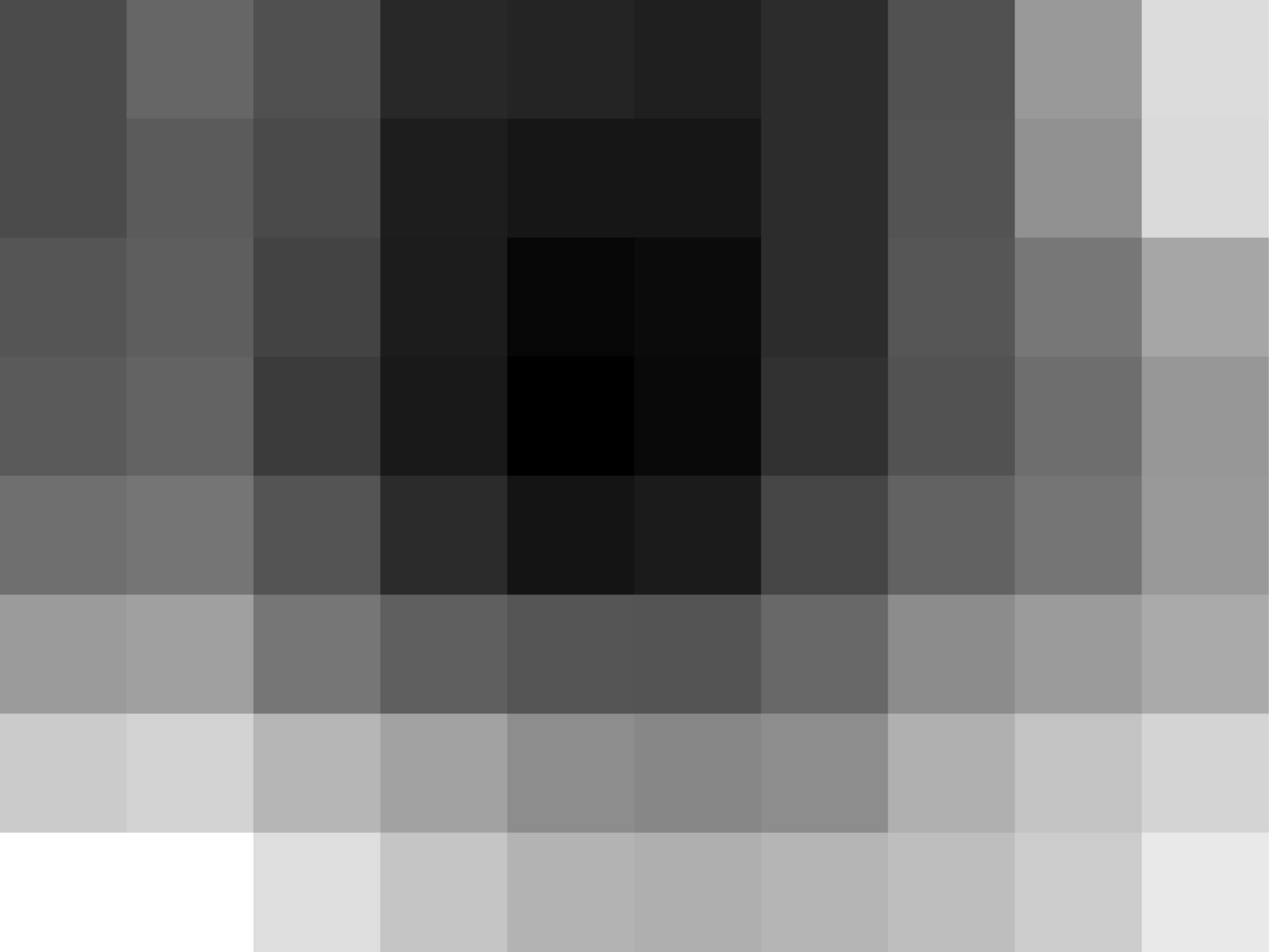}\\
(a) & (b)\\
\includegraphics[width=0.22\textwidth]{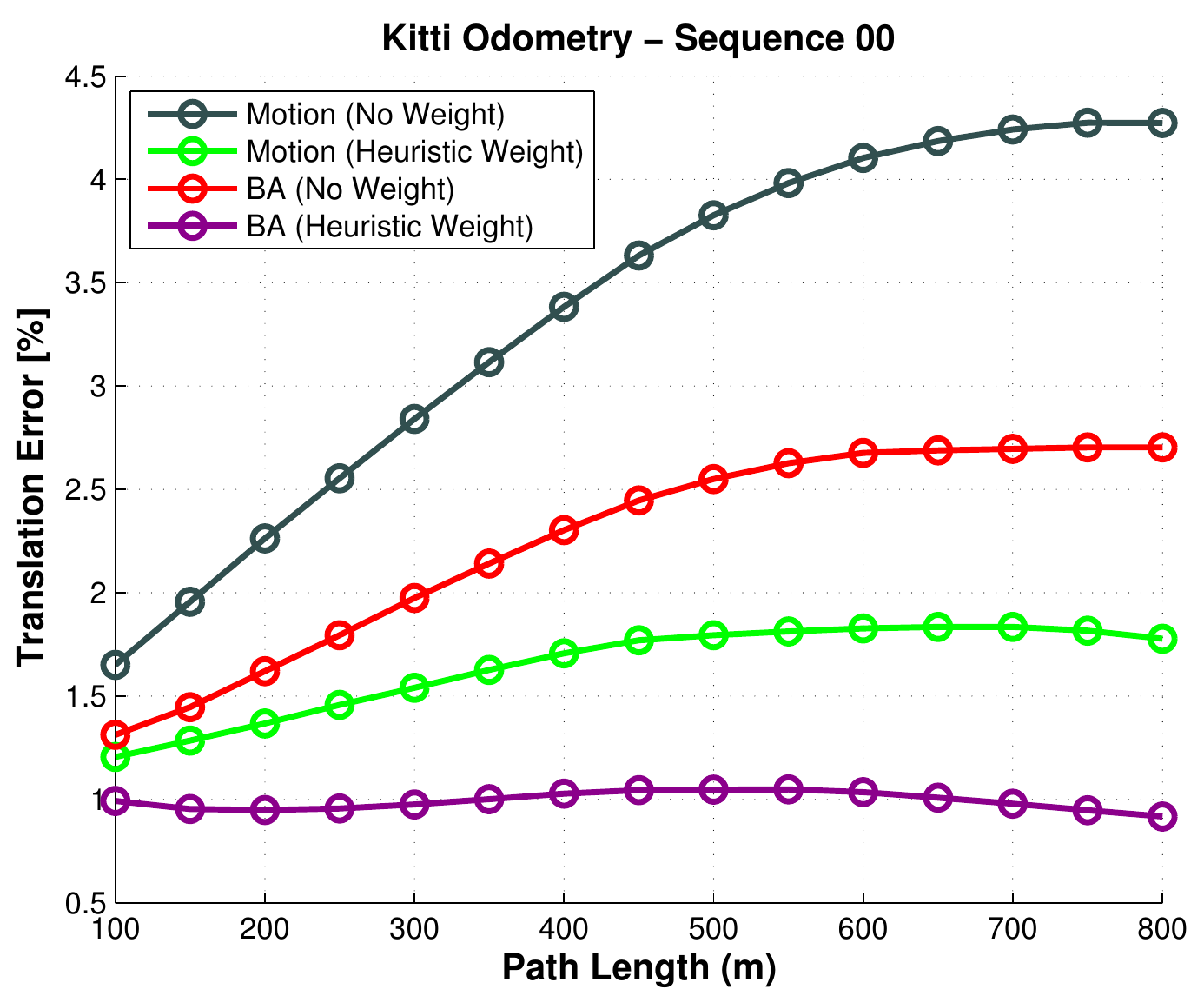}&
\includegraphics[width=0.22\textwidth]{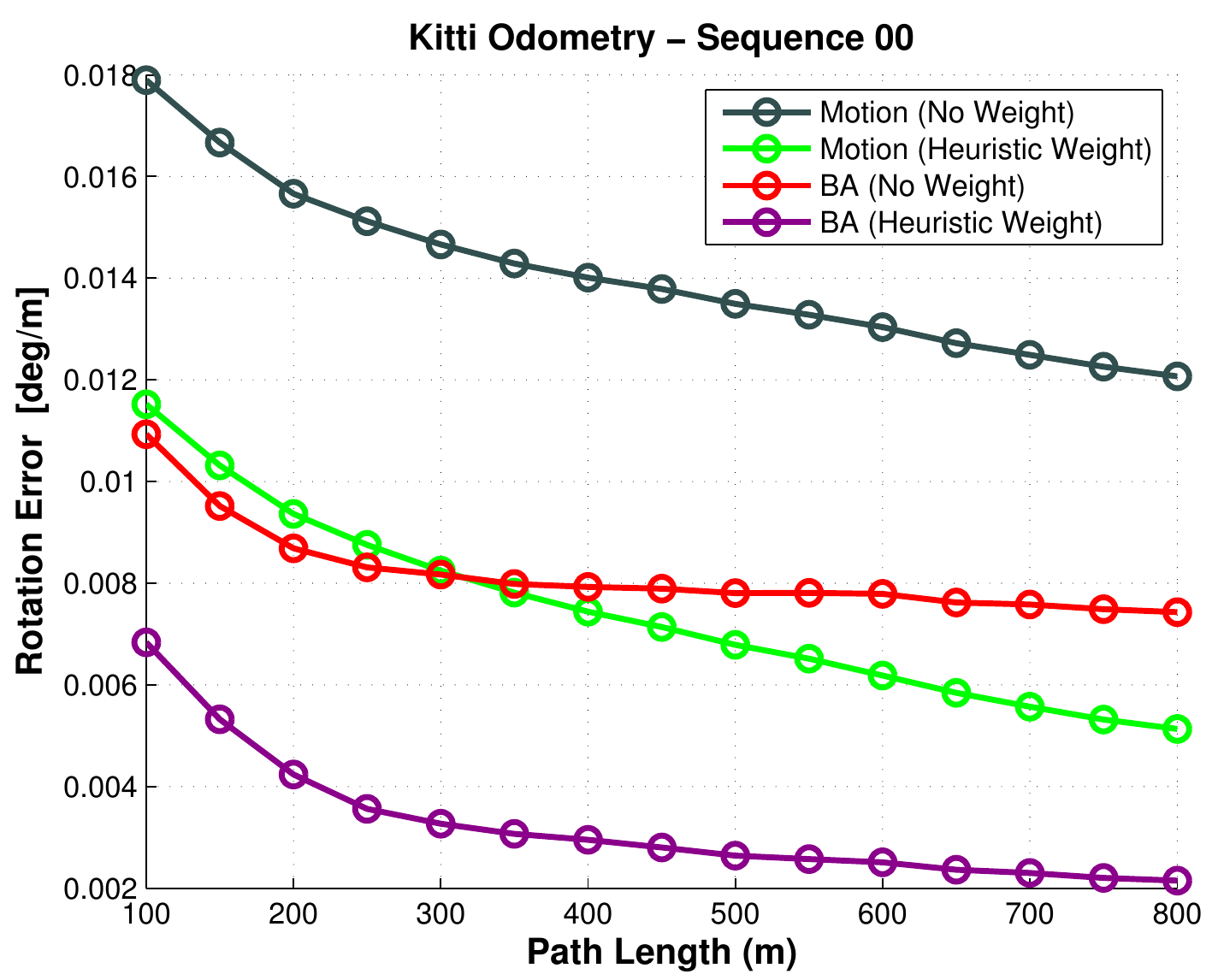}\\
(c) & (d)\\
\end{tabular}
\caption{Effect of feature weighting in the KITTI Odometry benchmark for the sequence 00. (a) Trajectories (b) Reprojection error 
histogram for the left camera (c) Average translation errors (d) Average rotation errors.}
\label{fig:weighting_bias}
\end{figure}

Although not mentioned in~\cite{Geiger11iv}, the LIBVISO2 library includes a heuristic feature weighting scheme with the aim of adding some 
robustness against calibration errors. Each stereo measurement has an associated scalar weight $w_i$ defined as:
\begin{equation} \label{eq:viso_weight}
w_{i} = \left(|u_{L} - u_{0}| / u_{0} + 0.05 \right)^{-1}
\end{equation}
where $u_{0}$ is the camera horizontal principal point. Those measurements that are closer to the principal point have a higher weight in 
the optimization, whereas measurements closer to image boundaries have lower weights since they are prone to be more erroneous due to 
calibration errors. Fig.~\ref{fig:weighting_bias} shows the effect of this weighting scheme in the camera motion accuracy, considering two types of refinement: motion only 
and BA.

As can be observed in Fig.~\ref{fig:weighting_bias}(c-d), there is a very significant improvement in accuracy 
when using the heuristic weighting scheme. An explanation of this improvement in accuracy is because the 
reprojection error distribution varies across the stereo image pair in the KITTI dataset~\cite{Kresso15visapp}, possibly due to poor camera calibration.
Our experiments corroborate this, revealing that there 
is a reprojection error bias when using the ground truth camera motion, which tends to be stronger for
observations closer to the image borders, as shown in Fig.~\ref{fig:weighting_bias}(b). On the other hand, reprojection errors are uniformly 
distributed in the New Tsukuba Stereo dataset, as expected since this is a synthetic dataset. In the rest of our experiments on the 
KITTI dataset, we will use the heuristic feature weighting scheme.

\subsection{Initialization}\label{sec:robust_results}
\begin{figure}[t]
\centering
\begin{tabular}{cc}
\includegraphics[width=0.22\textwidth]{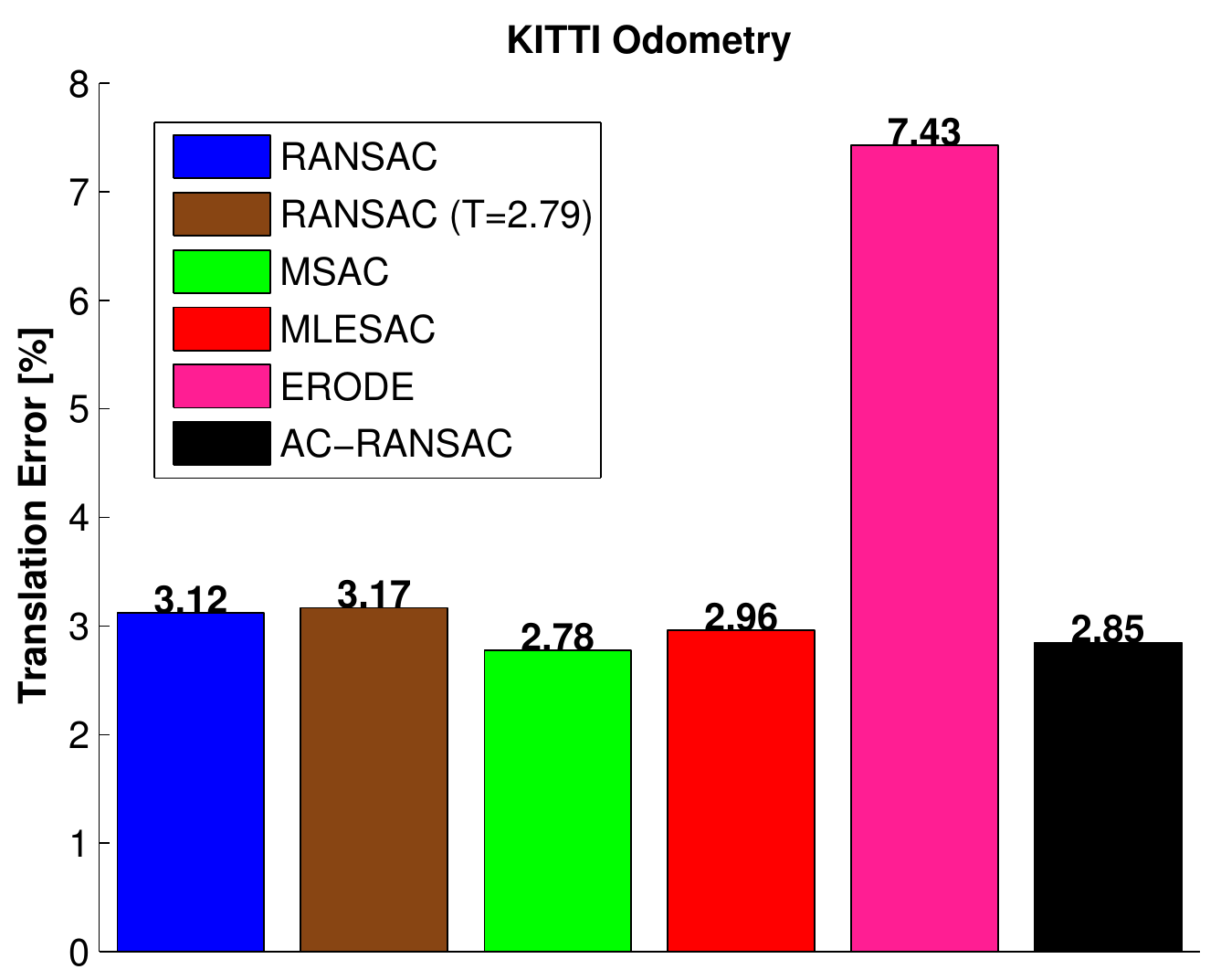}&
\includegraphics[width=0.22\textwidth]{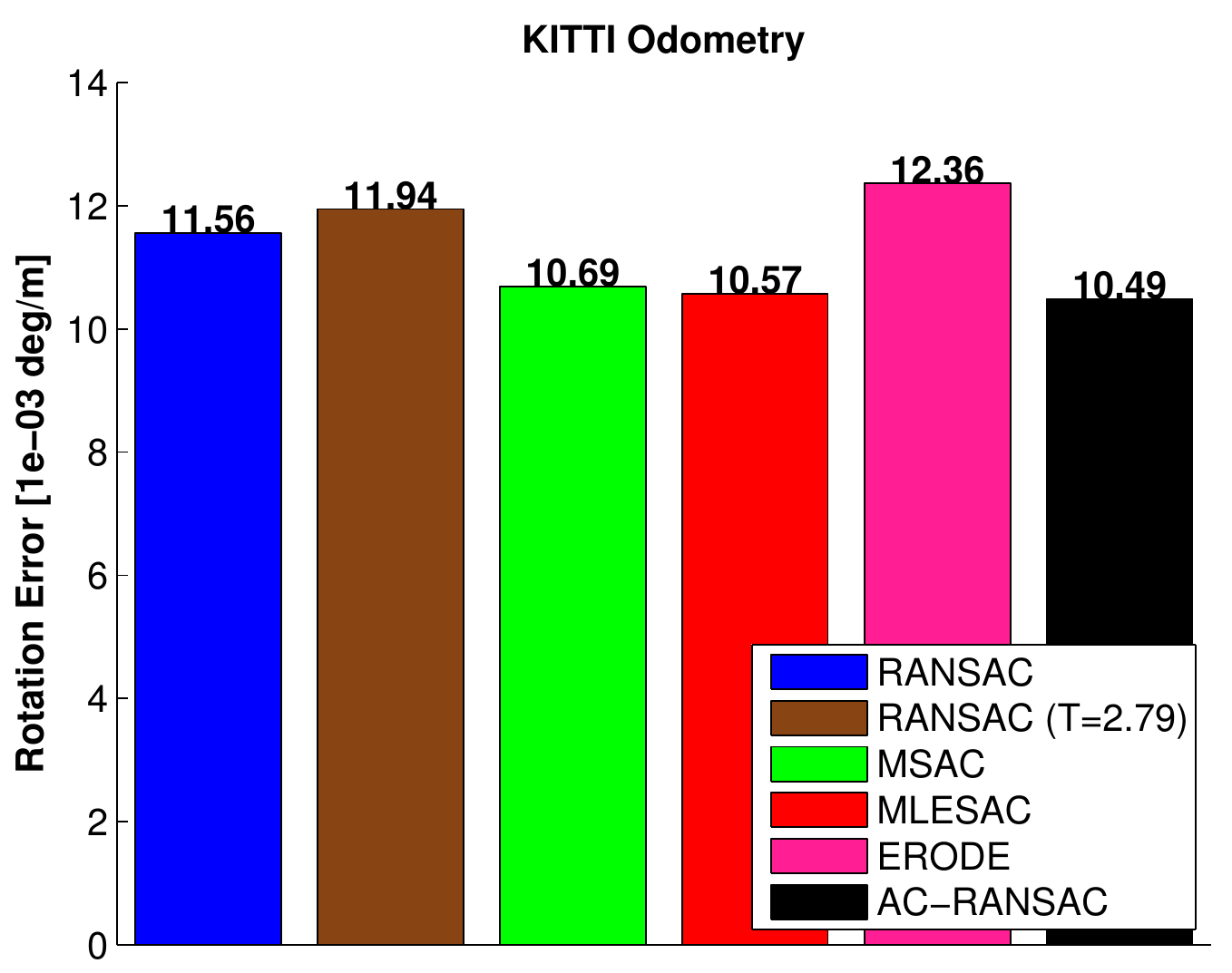}\\
(a) & (b) \\
\includegraphics[width=0.22\textwidth]{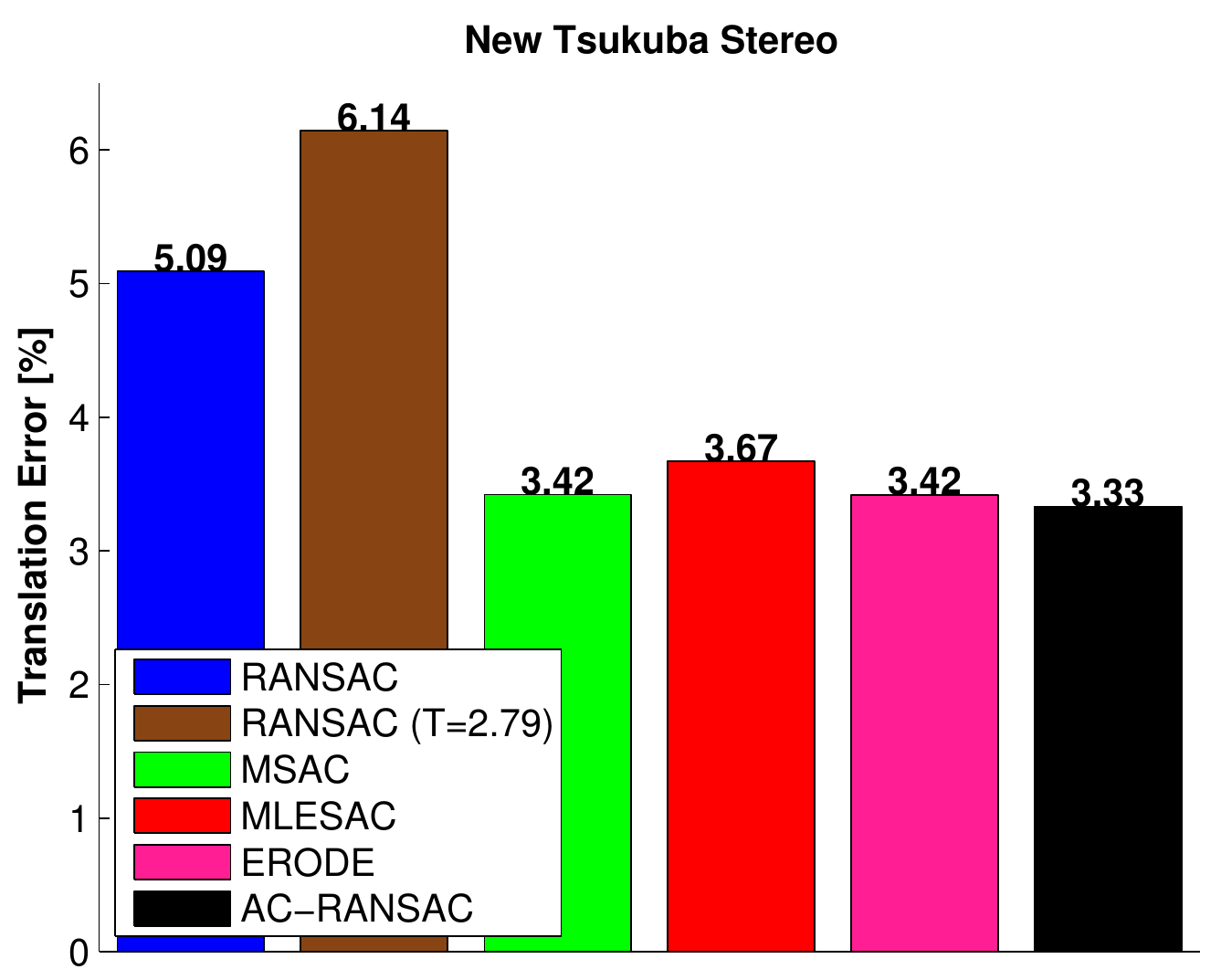}&
\includegraphics[width=0.22\textwidth]{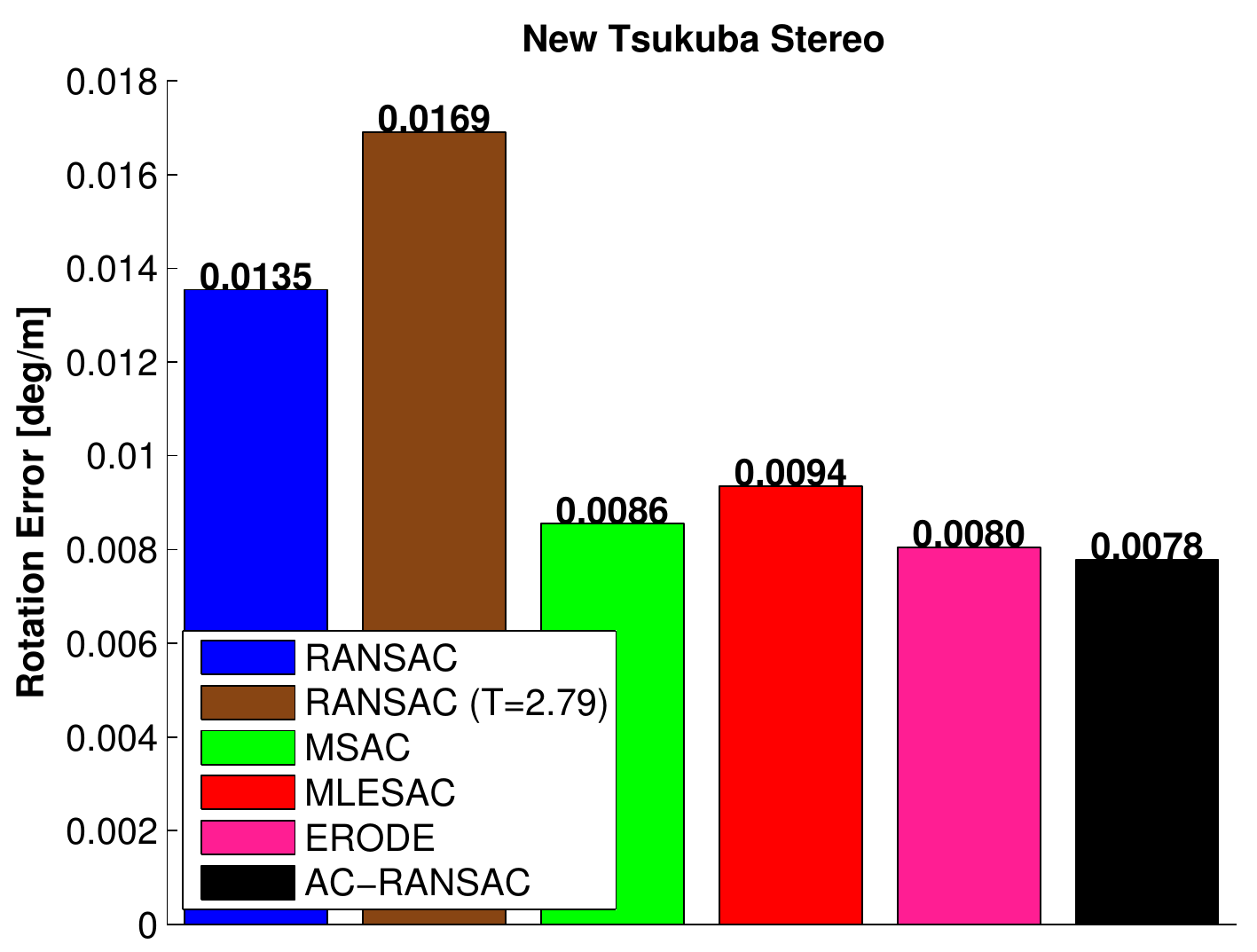}\\
(c) & (d) \\
\end{tabular}
\caption{Robust initialization results. Top: Average errors in translation (a) and rotation (b) for the KITTI Odometry training 
sequences (00-10) Bottom: Average errors in translation (c) and rotation (d) for the and New Tsukuba Stereo dataset, 
\textit{daylight} and \textit{fluorescent} illumination settings.}
\label{fig:initialization_eval}
\end{figure}

We performed 1000 iterations for RANSAC, MSAC, MLESAC and AC-RANSAC using the same set of minimal correspondences at each iteration. 
Fig.~\ref{fig:initialization_eval} depicts average translation and rotation errors for the KITTI training sequences and 
for the New Tsukuba Stereo dataset. 

According to the results in Fig.~\ref{fig:initialization_eval}, MSAC and AC-RANSAC are the two best performing methods. In the KITTI 
dataset, MSAC is slightly better in translation (2.78$\%$ versus 2.85$\%$) but AC-RANSAC is better 
in rotation (10.49\text{\sc{e}-03} deg/m versus 10.69\text{\sc{e}-03} deg/m). In the New Tsukuba Stereo dataset, AC-RANSAC 
outperforms MSAC and all the other methods both in translation and rotation. ERODE obtains bad performance in the KITTI 
dataset. The reason is because ERODE fails to compute a good camera motion when the motion between consecutive frames 
is large, as the initialization is far from the real camera motion and the cost function is multi-modal. In these scenarios, consensus based methods such as 
RANSAC are necessary in order to generate a good camera motion hypothesis. On the other hand, ERODE performs well in 
the New Tsukuba Stereo dataset, where the motion between frames is smaller than in the KITTI dataset.

\begin{figure}[t]
\centering
\begin{tabular}{cc}
\includegraphics[width=0.22\textwidth]{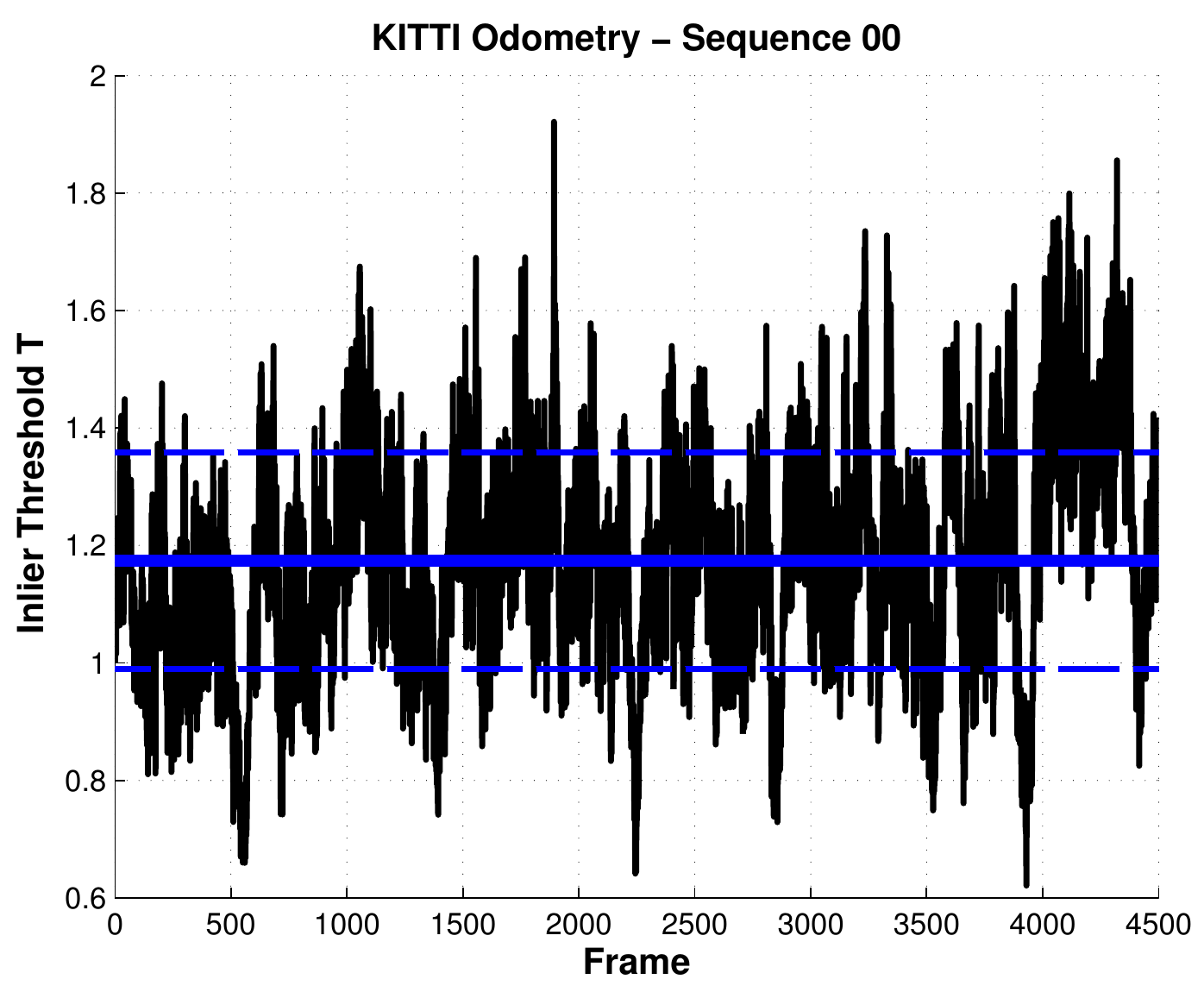}&
\includegraphics[width=0.22\textwidth]{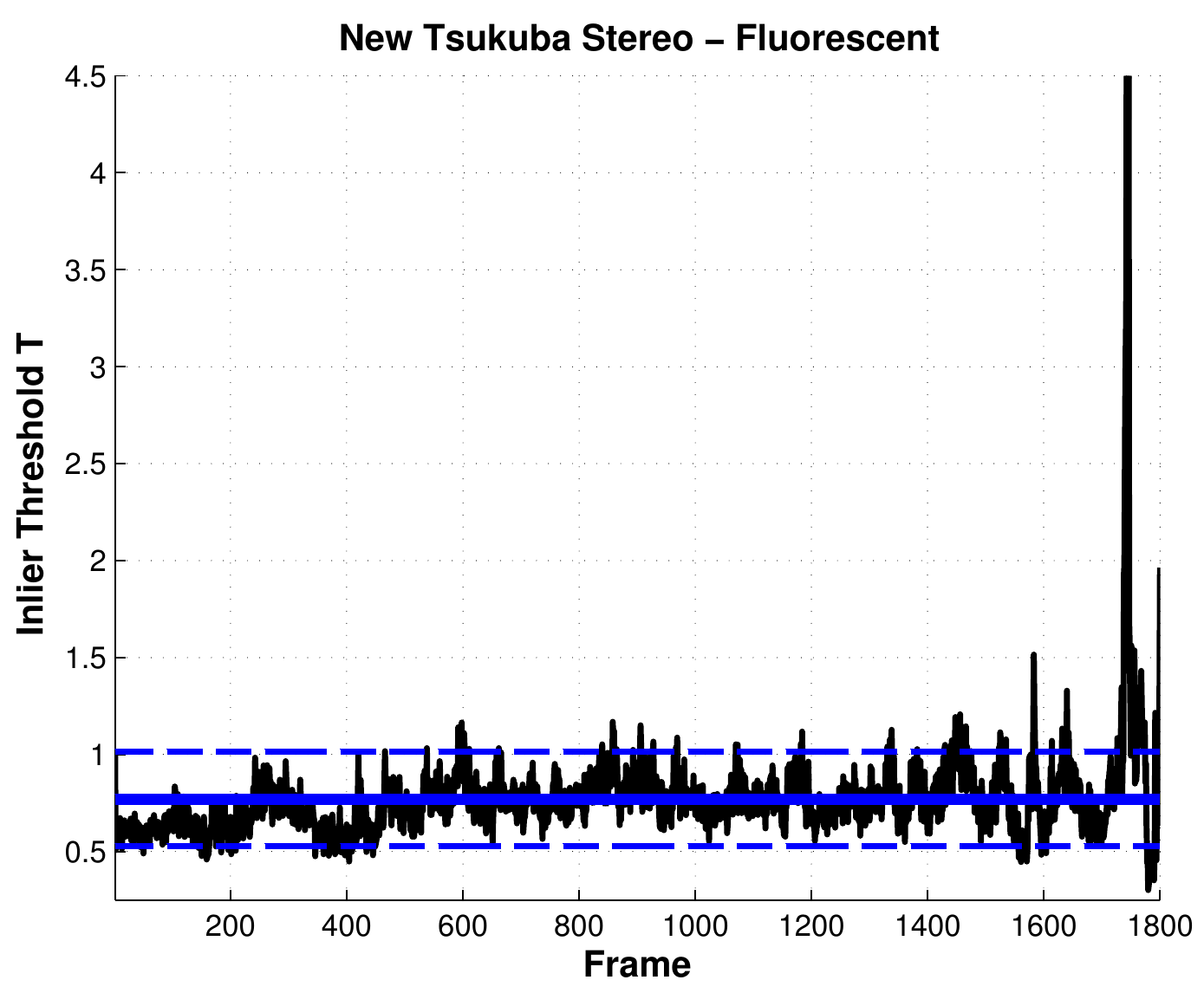}\\
(a) & (b)\\
\end{tabular}
\caption{Adaptive threshold $\thresh$ returned by AC-RANSAC per frame: (a) KITTI Odometry - Sequence 00 (b) New Tsukuba Stereo - 
Fluorescent. The solid blue line denotes the mean threshold for the sequence, while the discontinuous blue lines show the mean plus/minus 
the standard deviation of the threshold.}
\label{fig:ac_ransac_thresholds}
\end{figure}

Fig.~\ref{fig:ac_ransac_thresholds} depicts the adaptive thresholds per frame returned by AC-RANSAC for the first 
sequence from the KITTI odometry benchmark and the \textit{fluorescent} sequence from the New Tsukuba Stereo dataset. 
As has been shown in Fig.~\ref{fig:initialization_eval}, the choice of a proper threshold value $\thresh$ is 
critical in order to get good accuracy. This value may vary between different datasets and between different image 
pairs within the same sequence. AC-RANSAC solves this problem by finding an adaptive threshold for each pair of frames.
%

\subsection{Refinement}\label{sec:refinement_results}
\begin{figure}[t]
\centering
\begin{tabular}{c}
\includegraphics[width=0.45\textwidth]{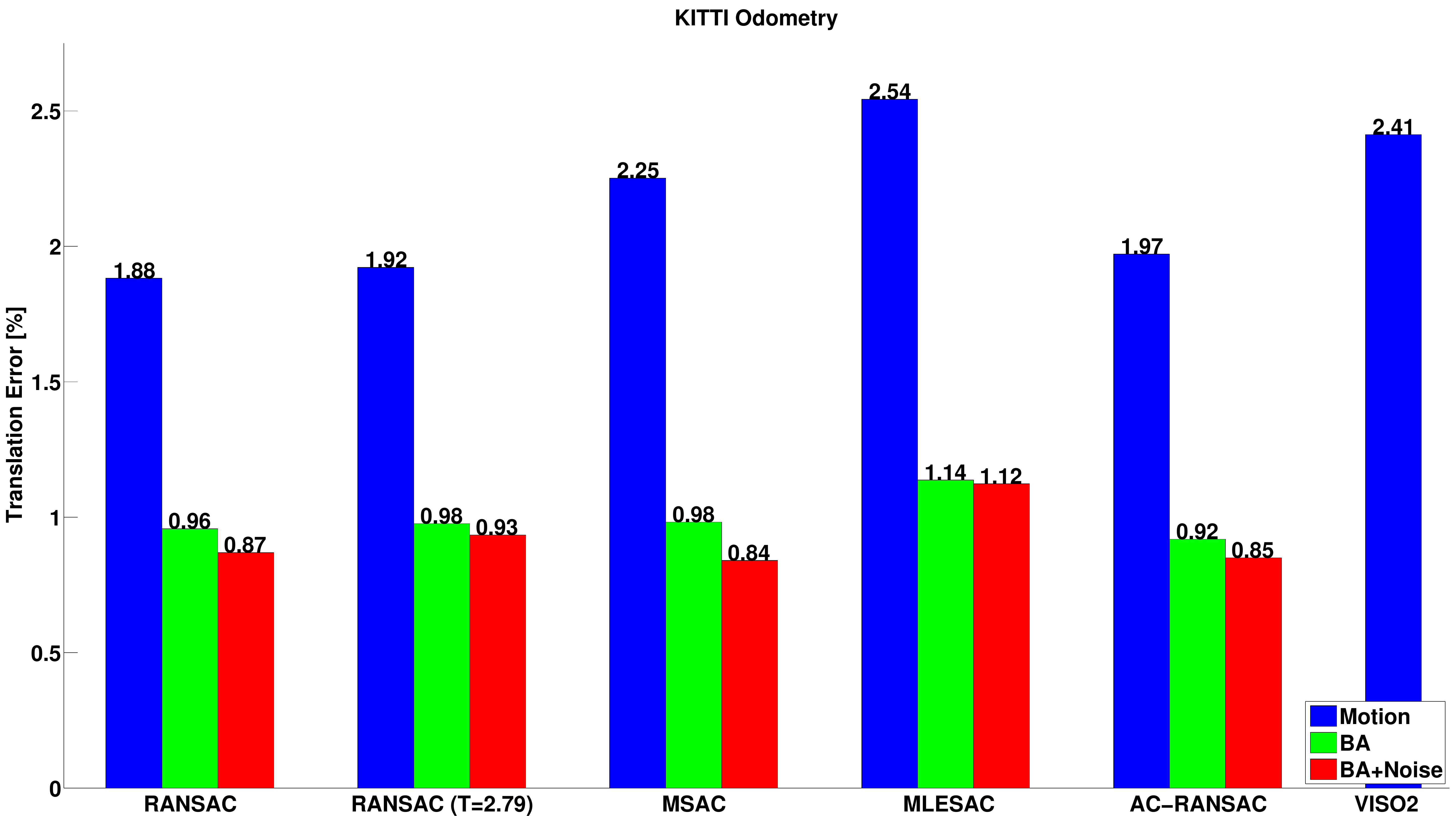} \\
(a) \\
\includegraphics[width=0.45\textwidth]{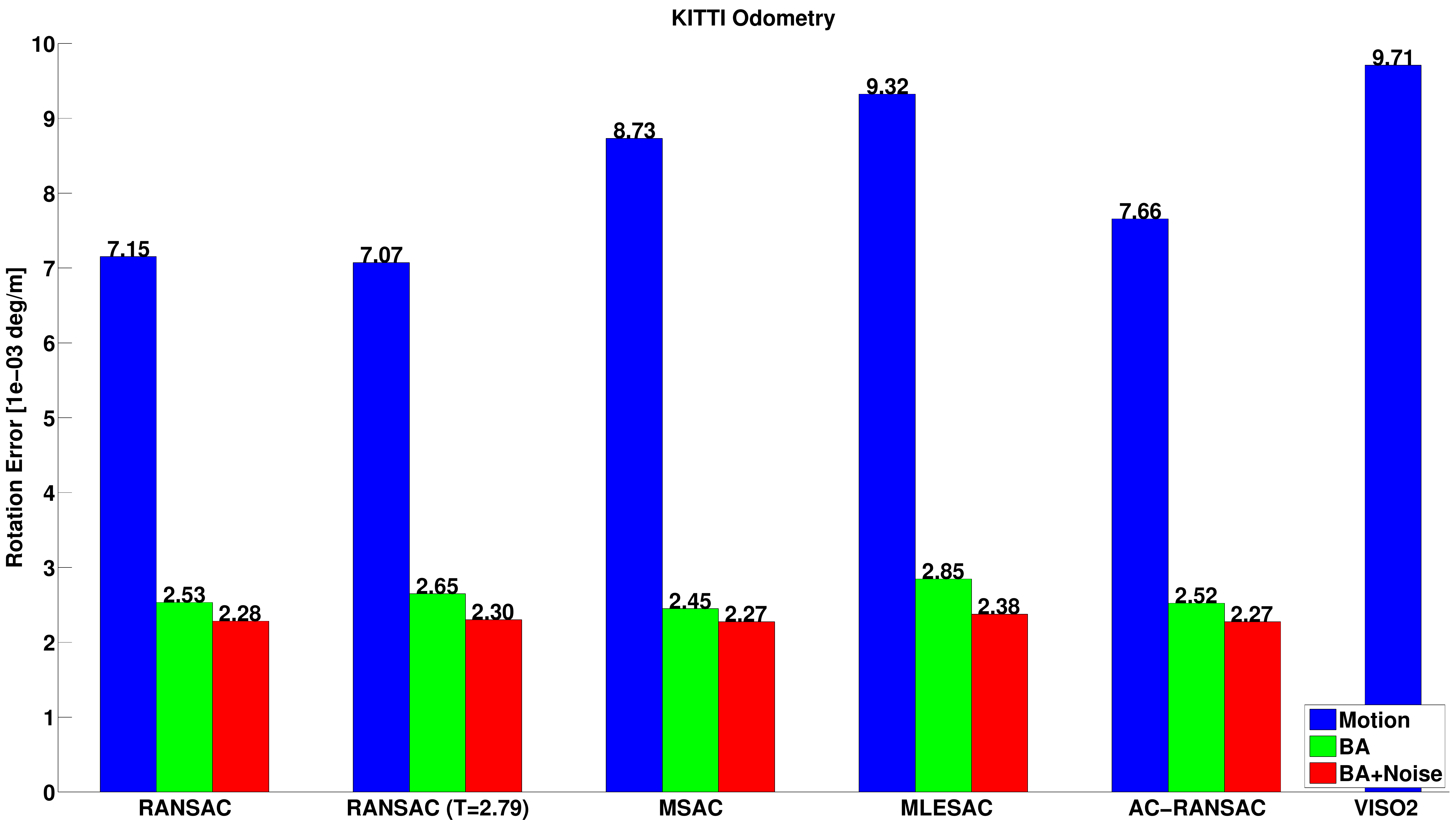} \\
(b)
\end{tabular}
\caption{Average errors in translation (a) and rotation (b) for the KITTI odometry training sequences (00-10), considering three different 
refinement strategies: motion only, BA and BA plus inlier noise distribution.}
\label{fig:kitti_refinement_eval}
\end{figure}

Fig.~\ref{fig:kitti_refinement_eval} depicts average translation and rotation errors for the KITTI odometry training sequences, considering 
the different model refinement strategies (motion, BA, BA plus inlier noise distribution). We do not show results for ERODE in the 
refinement evaluation for the KITTI dataset since this method fails to obtain good initial results. Best results are again obtained with 
MSAC and AC-RANSAC with BA plus inlier noise distribution refinement, giving errors of 0.84$\%$ in 
translation and 2.27\text{\sc{e}-03} deg/m in rotation. 

Table~\ref{tab:kitti_refinement_improvements} shows the improvements in camera motion accuracy by considering the BA and the BA 
plus inlier noise distribution refinement strategies \wrt the motion only refinement. The improvement in $\%$ is computed as 
$100\cdot((A_{motion}/A_{ref^*})-1)$, where $A_{motion}$ and $A_{ref^*}$ denote the accuracy of the motion only refinement and BA or BA 
plus inlier noise distribution strategies respectively. BA, \ie including 3D structure in the refinement optimization, considerably improves the camera motion accuracy for all the cases. 
In addition, BA plus inlier noise distribution also helps to further improve the camera motion accuracy. 
\begin{table}[htpb]
  \begin{small}
  \begin{center}
    \begin{tabular}{|c|c|c|c|}
      \hline
       \multicolumn{2}{|c|}{} & \textbf{BA} &  \textbf{BA+Inlier Noise}\\
      \hline
      \textbf{RANSAC} & Translation & 95.83 \% & 116.09 \%\\
      & Rotation & 182.61 \% & 213.59 \%\\
      \hline
      \textbf{RANSAC} & Translation & 95.91 \% & 106.45 \%\\
      T = 2.79 & Rotation & 166.79 \% & 207.39 \%\\
      \hline
      \textbf{MSAC} & Translation & 129.59 \% & 167.85 \%\\
       & Rotation & 256.32 \% & 284.58 \%\\
      \hline
      \textbf{MLESAC} & Translation & 122.81 \% & 126.78 \% \\
       & Rotation & 227.01 \% & 291.59 \%\\
      \hline
      \textbf{AC-RANSAC} & Translation & 114.13 \% & 131.76 \%\\
       & Rotation & 203.96 \% & 237.45 \%\\
      \hline
    \end{tabular}
  \end{center}
  \end{small}
  \caption{Model refinement improvements in the KITTI Odometry training sequences (00-10) \wrt motion only 
refinement.}\label{tab:kitti_refinement_improvements}
\end{table}

\begin{figure}[t]
\centering
\begin{tabular}{c}
\includegraphics[width=0.45\textwidth]{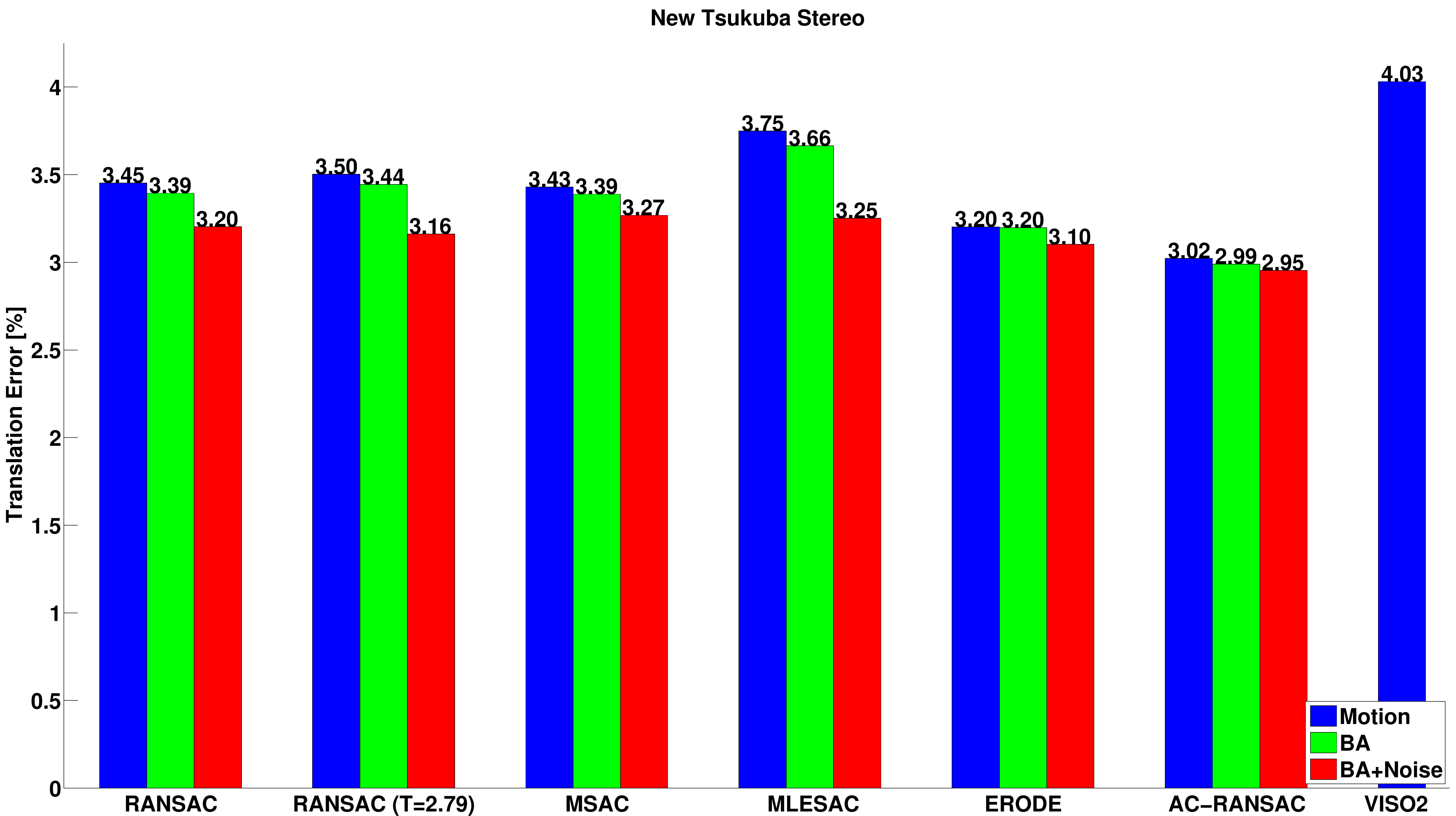} \\
(a) \\
\includegraphics[width=0.45\textwidth]{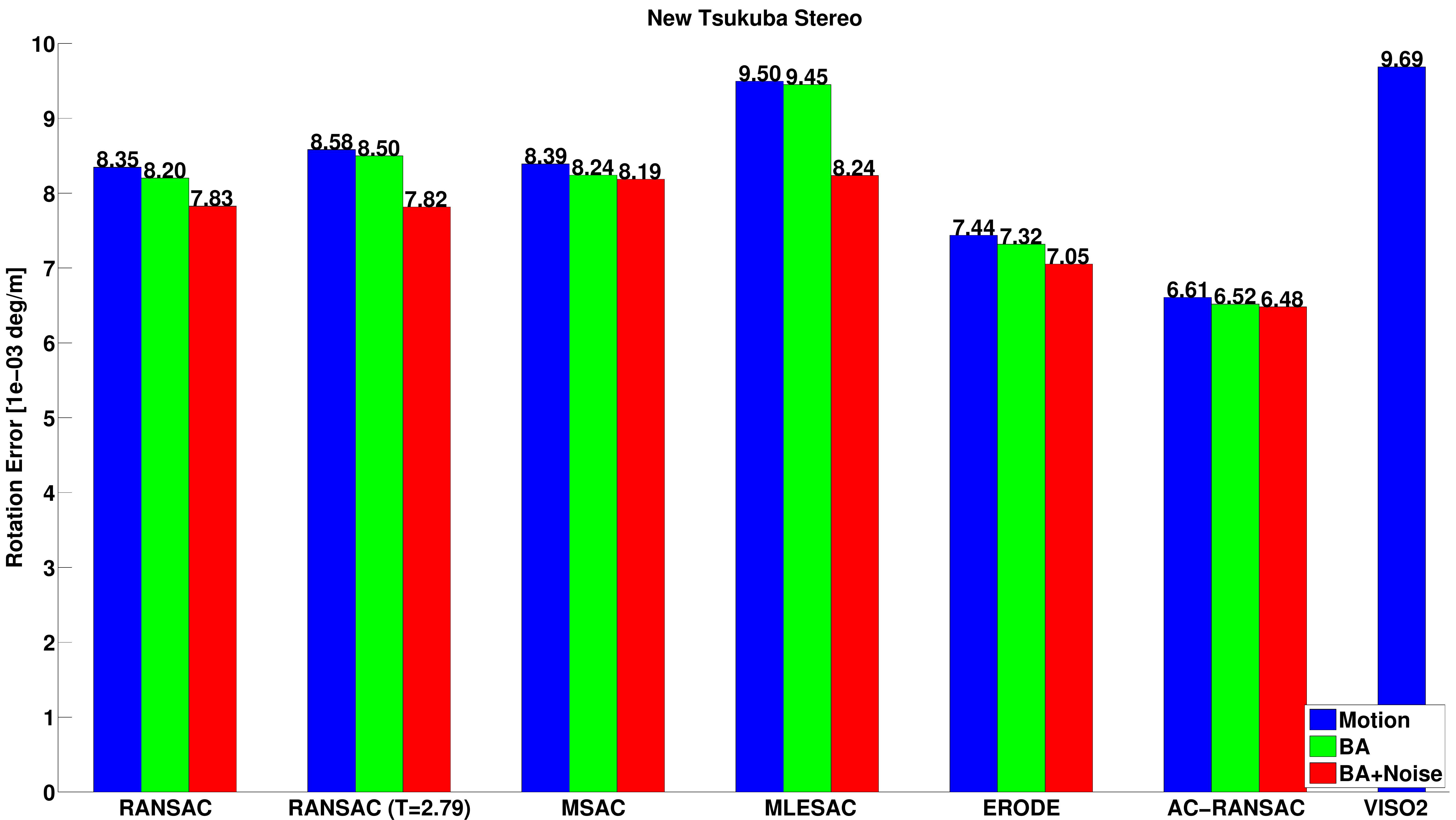} \\
(b) \\
\end{tabular}
\caption{Average errors for the New Tsukuba Stereo dataset, \textit{daylight} and \textit{fluorescent} sequences. (a) Translation error (b) 
Rotation error.}
\label{fig:tsukuba_refinement_eval}
\end{figure}

Fig.~\ref{fig:tsukuba_refinement_eval} depicts average translation and rotation errors for the New Tsukuba Stereo \textit{daylight} and 
\textit{fluorescent} sequences. In this case, AC-RANSAC alongside BA plus inlier noise distribution refinement is the best performing 
method with errors of 2.95$\%$ in translation and 6.48\text{\sc{e}-03} deg/m.

Table~\ref{tab:tsukuba_refinement_improvements} shows the improvements in camera motion accuracy by considering the BA 
and the BA plus inlier noise distribution refinement strategies \wrt the motion only refinement in the New Tsukuba 
Stereo dataset. While significant, the improvements due to the refinement of 3D structure and inlier noise 
distribution are much larger in the KITTI dataset than in the New Tsukuba Stereo. The main explanation is that KITTI is 
a real dataset and therefore calibration is more prone to errors than in the New Tsukuba dataset, which is a synthetic 
dataset. Therefore, image observations and 3D points are noisier in KITTI than in the New Tsukuba Stereo dataset.
\begin{table}[htpb]
  \begin{small}
  \begin{center}
    \begin{tabular}{|c|c|c|c|}
      \hline
       \multicolumn{2}{|c|}{} & \textbf{BA} &  \textbf{BA+Inlier Noise}\\
      \hline
      \textbf{RANSAC} & Translation & 1.77 \% & 7.81 \%\\
      & Rotation & 1.83 \% & 6.64 \%\\
      \hline
      \textbf{RANSAC} & Translation & 1.74 \% & 10.76 \%\\
      T = 2.79 & Rotation & 0.94 \% & 9.72 \%\\
      \hline
      \textbf{MSAC} & Translation & 1.18 \% & 4.89 \%\\
       & Rotation & 1.82 \% & 2.44 \%\\
      \hline
      \textbf{MLESAC} & Translation & 3.01 \% & 16.00 \% \\
       & Rotation & 0.53 \% & 15.29 \%\\
      \hline
      \textbf{ERODE} & Translation & 0.00 \% & 3.22 \% \\
       & Rotation & 1.63 \% & 5.53 \%\\
      \hline
      \textbf{AC-RANSAC} & Translation & 1.00 \% & 2.37 \%\\
       & Rotation & 1.38 \% & 2.01 \%\\
      \hline
    \end{tabular}
  \end{center}
  \end{small}
  \caption{Model refinement improvements in the New Tsukuba Stereo dataset \wrt motion only 
refinement.}\label{tab:tsukuba_refinement_improvements}
\end{table}

\begin{figure}[t]
\centering
\begin{tabular}{c}
\includegraphics[width=0.4\textwidth]{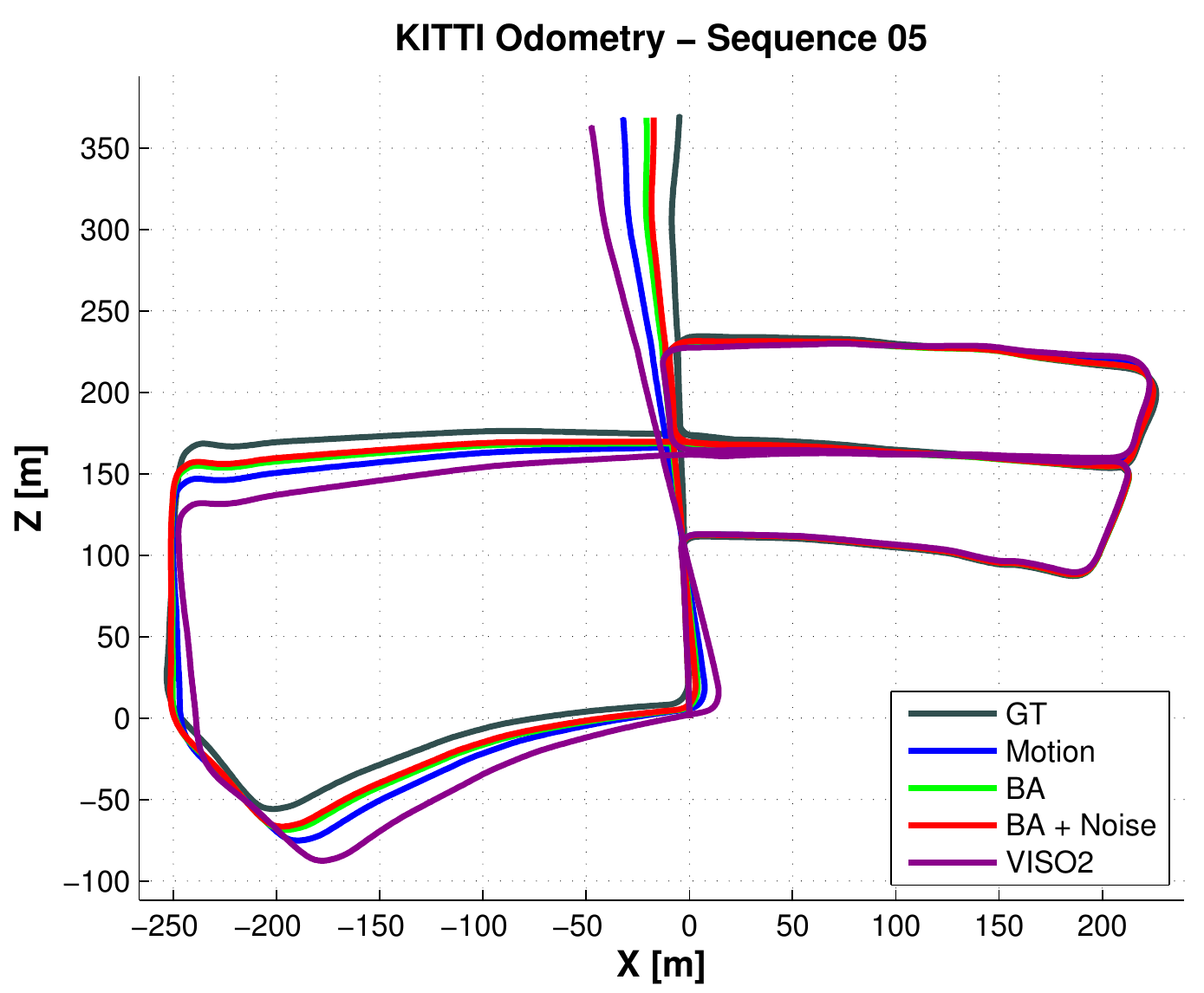}\\
(a) \\
\includegraphics[width=0.4\textwidth]{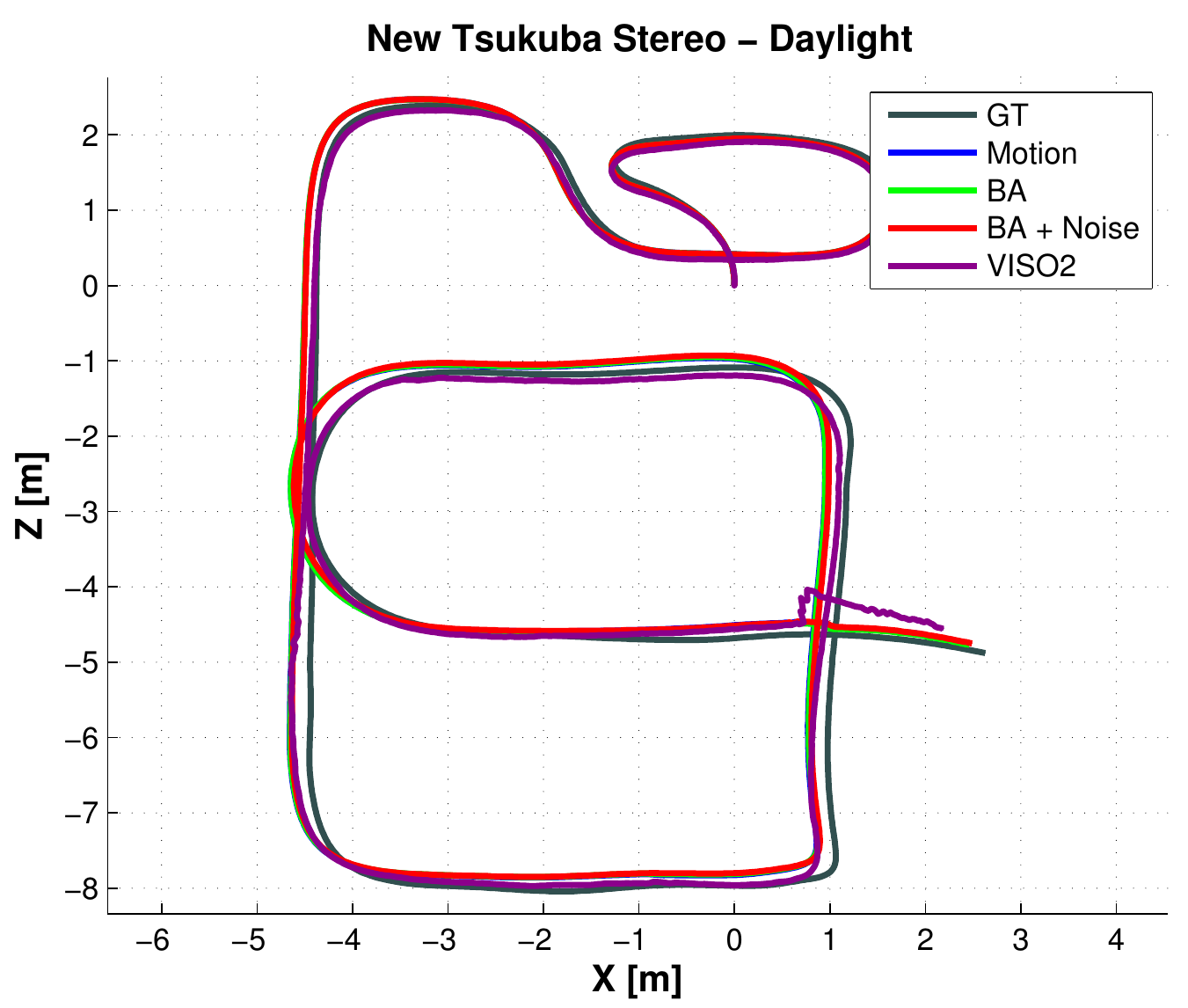} \\
(b)
\end{tabular}
\caption{Reconstructed trajectories considering AC-RANSAC alongside refinement strategies and VISO2: (a) KITTI Odometry - Sequence 05 (b) 
New Tsukuba Stereo - Daylight.}
\label{trajectories_comparison}
\end{figure}

We performed the same evaluation using the LIBVISO2 library. LIBVISO2 obtained a mean average error in translation of 
2.41$\%$ and 9.71\text{\sc{e}-03} deg/m in rotation for the KITTI Odometry training dataset. Regarding the experiments in the 
New Tsukuba Stereo dataset, LIBVISO2 obtained a mean average error in translation 4.03$\%$ of and 9.69\text{\sc{e}-03} 
deg/m in rotation. Fig.~\ref{trajectories_comparison} shows an example of the estimated trajectories 
using AC-RANSAC and the different refinement strategies \wrt to LIBVISO2 for the KITTI Odometry Sequence 05 and for the New Tsukuba 
Stereo daylight setting. According to the results, our different refinement techniques in our stereo VO evaluation greatly outperform the 
results of LIBVISO2 in the two analyzed datasets.
%
\subsection{Timing Evaluation}
\begin{figure}[t]
\centering
\begin{tabular}{cc}
\includegraphics[width=0.22\textwidth]{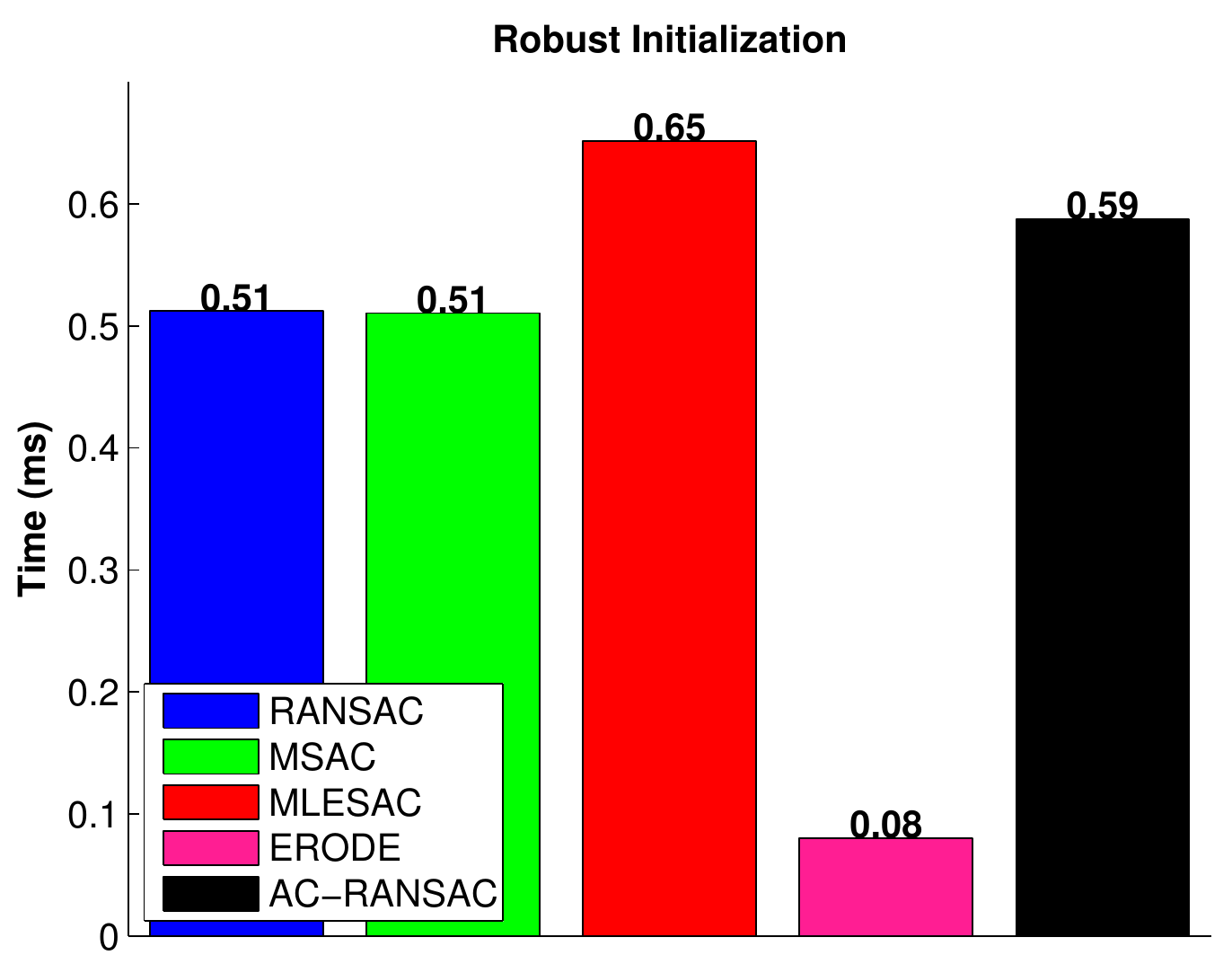} &
\includegraphics[width=0.22\textwidth]{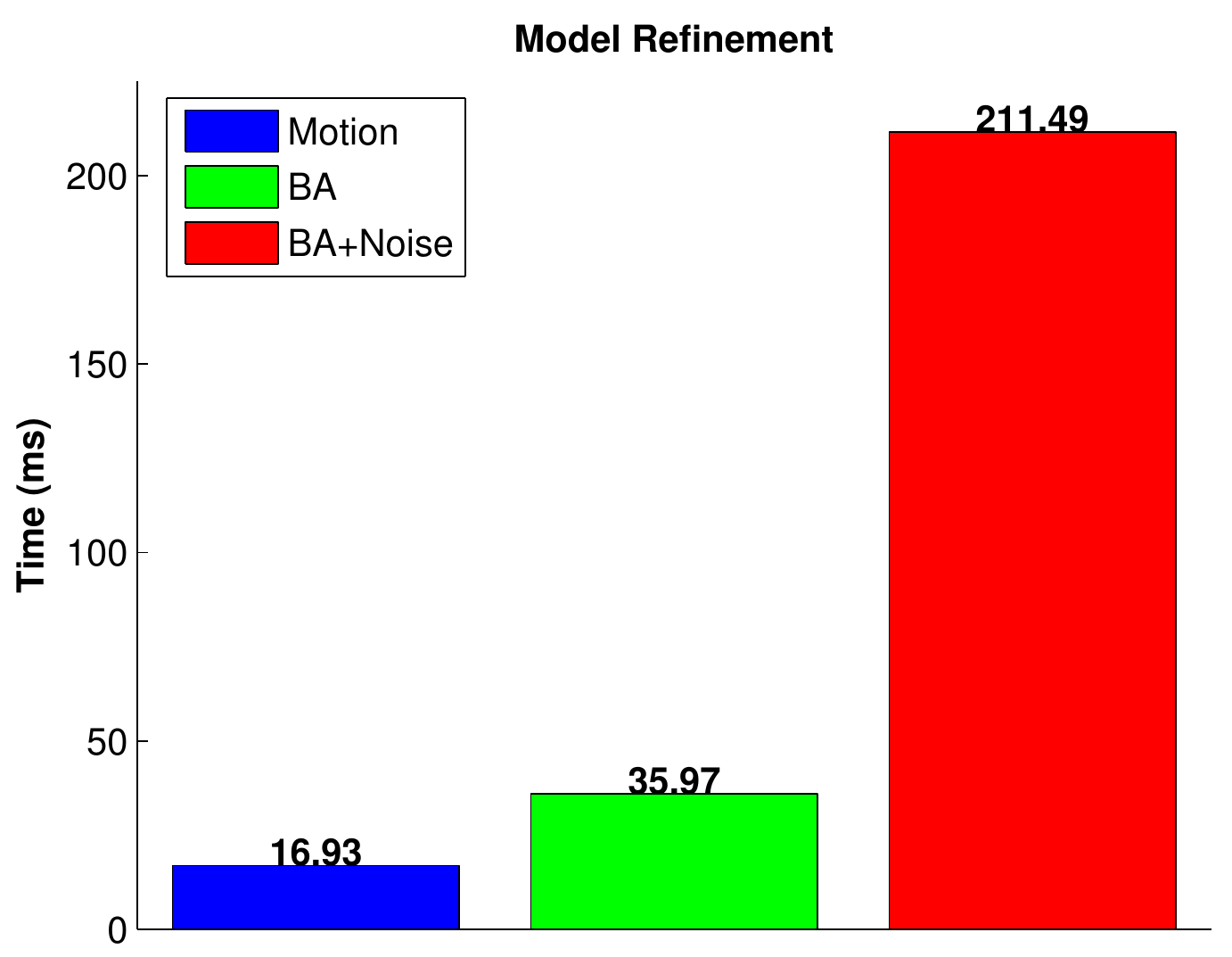}\\
(a) & (b)
\end{tabular}
\caption{(a) Robust initialization timing results: average computation time in ms per RANSAC iteration. (b) Model refinement timing 
results: average computation time in ms for each optimization strategy. In both cases, 1000 putative correspondences were considered. Timing 
results were obtained using a single threaded implementation on a 3.00GHz desktop computer.}
\label{fig:timing}
\end{figure}

Fig.~\ref{fig:timing}(a) depicts a timing evaluation for the robust initialization and refinement strategies. ERODE is 
the fastest method. MLESAC is the slowest method since the inlier probability, $\inlier_ratio$, needs to be 
estimated per camera motion hypothesis using the EM algorithm. AC-RANSAC is also expensive since it needs to sort the residuals per hypothesis. 
RANSAC and MSAC obtain similar timing results. Note that it is possible to use other acceleration schemes to speed-up 
consensus methods such as~\cite{Chum04cvpr}. In the case of AC-RANSAC, the acceleration scheme ORSA~\cite{Moisan04ijcv} 
speeds-up the robust initialization since once it finds a good hypothesis, the procedure is restarted for a reduced 
number of iterations, drawing new samples from the best inlier set found so far. 

Fig.~\ref{fig:timing}(b) depicts timing results for the refinement strategies, where the refinement of motion, structure and inlier noise 
distribution strategy is the most computationally expensive strategy. 
%

\section{Conclusions}\label{sec:conclusions}
In this paper we have performed an evaluation of image feature localisation noise models, alongside optimization 
strategies, in the context of feature-based stereo VO. Our evaluation shows that noise models that are more adaptable to the varying nature of the noise generally 
perform better. Despite the bias in the KITTI Odometry evaluation, our results are confirmed using the New Tsukuba Stereo dataset.
In addition, we believe this paper presents the first use of AC-RANSAC, and also the first joint optimization of BA and inlier noise distribution parameters, in stereo VO.

%
%
\bibliographystyle{IEEEtran}
\bibliography{references}
%

\end{document}